\documentclass[pdflatex,sn-mathphys-num]{sn-jnl}

\makeatletter
\newcommand{\pushright}[1]{\ifmeasuring@#1\else\omit\hfill$\displaystyle#1$\fi\ignorespaces}
\newcommand{\pushleft}[1]{\ifmeasuring@#1\else\omit$\displaystyle#1$\hfill\fi\ignorespaces}
\makeatother

\usepackage{booktabs}
\let\cline\cmidrule
\usepackage{siunitx}
\usepackage{bm}
\usepackage{caption}
\usepackage{amsmath}

\newcommand{\ind}{\text{\usefont{U}{bbold}{m}{n}1}} 

\usepackage{amssymb}

\usepackage[table]{xcolor}

\begin{document}

\title[Patient-level Information Extraction by Consistent Integration of Text and Tab. Evidence with BNs]{Patient-level Information Extraction by Consistent Integration of Textual and Tabular Evidence with Bayesian Networks}

\author*[1,2]{\fnm{Paloma} \sur{Rabaey}}\email{paloma.rabaey@ugent.be}
\equalcont{These authors contributed equally to this work.}

\author*[1,2]{\fnm{Adrick} \sur{Tench}}\email{adrick.tench@ugent.be}
\equalcont{These authors contributed equally to this work.}

\author[3]{\fnm{Stefan} \sur{Heytens}}

\author[1,2]{\fnm{Thomas} \sur{Demeester}}

\affil[1]{\orgdiv{Faculty of Engineering and Architecture}, \orgname{Ghent University}, \orgaddress{\city{Ghent},\country{Belgium}}}

\affil[2]{\orgname{imec}, \orgaddress{\city{Leuven}, \country{Belgium}}}

\affil[3]{\orgname{Ghent University Hospital}, \orgaddress{\city{Ghent}, \country{Belgium}}}

\abstract{
Electronic health records (EHRs) form an invaluable resource for training clinical decision support systems. To leverage the potential of such systems in high-risk applications, we need large, structured tabular datasets on which we can build transparent feature-based models. While part of the EHR already contains structured information (e.g. diagnosis codes, medications, and lab results), much of the information is contained within unstructured text (e.g. discharge summaries and nursing notes). In this work, we propose a method for multi-modal patient-level information extraction that leverages both the tabular features available in the patient’s EHR (using an expert-informed Bayesian network) as well as clinical notes describing the patient’s symptoms (using neural text classifiers). We propose the use of \emph{virtual evidence} augmented with a \emph{consistency node} to provide an interpretable, probabilistic fusion of the models' predictions. The consistency node improves the calibration of the final predictions compared to virtual evidence alone, allowing the Bayesian network to better adjust the neural classifier’s output to handle missing information and resolve contradictions between the tabular and text data. We show the potential of our method on the SimSUM dataset, a simulated benchmark linking tabular EHRs with clinical notes through expert knowledge.
}


\maketitle

\section{Introduction}


\begin{figure*}[t]
    \centering
    \includegraphics[width=\textwidth]{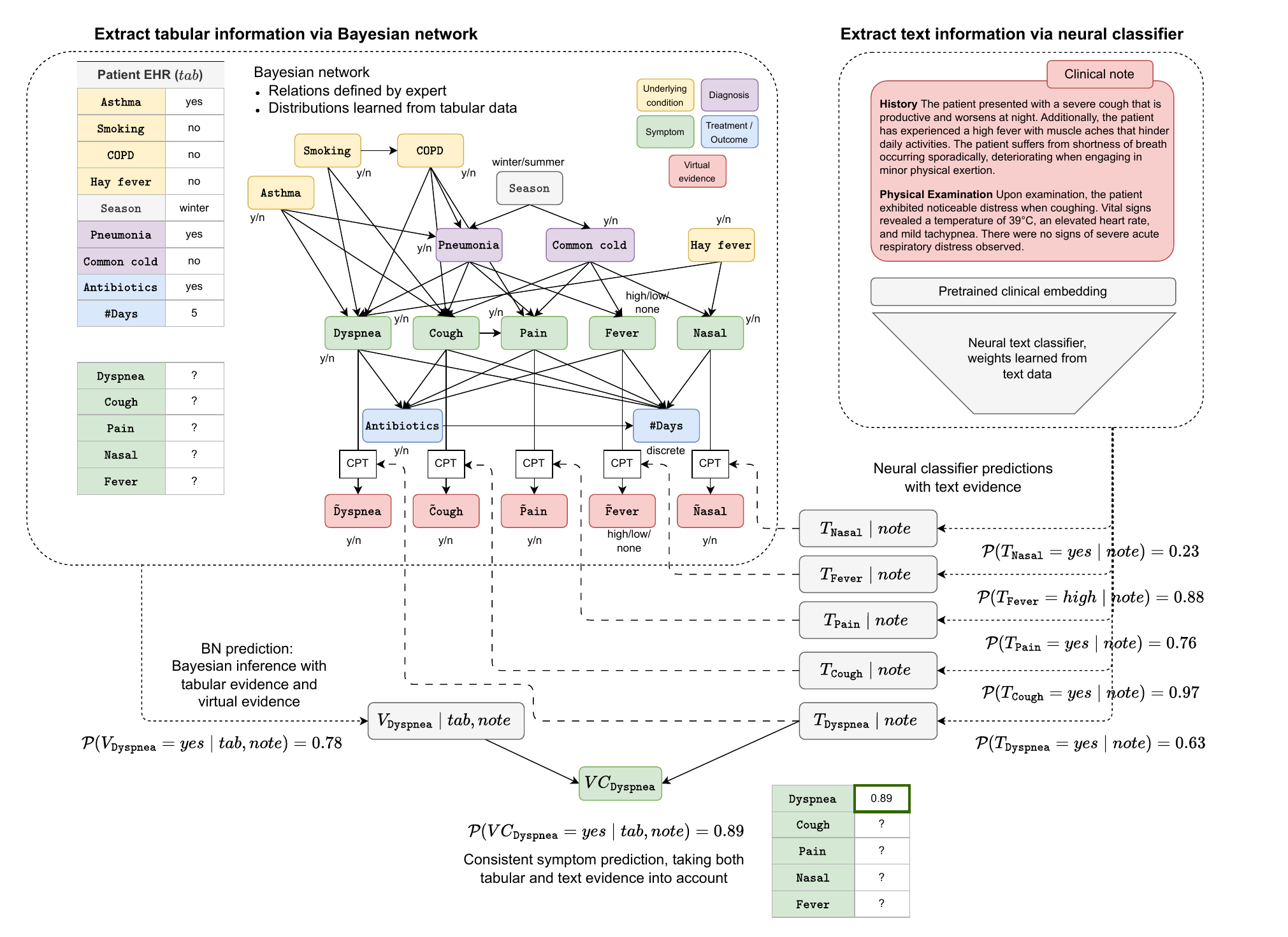}
    \caption{Overview of our patient-level information extraction method which integrates both tabular and text evidence. As an example, we show how to extract the probability that a patient suffers from \texttt{Dyspnea}, given tabular evidence that is already encoded in the EHR, a clinical note describing the patient's symptoms, and an expert-defined \textbf{Bayesian network (BN)} structure. On the right, the \textbf{neural classifiers} infer probabilities that the \textbf{text} mentions each symptom, with a 63\% confidence for \texttt{Dyspnea} (in this case, ``dyspnea'' is not mentioned verbatim). The classifiers' probabilities for each symptom are provided as \textbf{virtual evidence} to the \textbf{BN}, via the red ``virtual" nodes in the network. Given all \textbf{tabular} and \textbf{virtual evidence}, the \textbf{BN} infers that the patient has a 78\% chance of \texttt{Dyspnea} -- since the patient has both \texttt{Pneumonia} and \texttt{Asthma}, this probability is high. The consistency node $VC_{\texttt{Dyspnea}}$ combines these probabilities, arriving at an 89\% chance that this patient has \texttt{Dyspnea}. Part of this figure is adapted from \citet{SimSUM}.}
    \label{fig:consistency_overview}
\end{figure*}

In healthcare, storing patient information in a structured format is essential for ensuring continuity of care. This information, such as diagnosis codes, medication, and lab results, is usually stored in Electronic Health Records (EHRs). Often, these also contain a large quantity of unstructured free text, such as discharge summaries, nursing notes, procedural descriptions, and more \citep{extracting_information_EMR}. All this information forms an invaluable resource for training clinical decision support systems, which have the potential to partially automate care processes such as diagnosis and treatment planning \citep{CDS_infectious_diseases, medBERT, BEHRT}. While the recent rise in large language models has shown great potential for processing clinical notes \citep{clinicalT5, MedPalm, BioMistral}, these black-box systems lack interpretability \citep{black_box, explainability_LLMs, opportunities_challenges_biomedicine}. In high-risk clinical applications, one should prefer robust and transparent systems built on simpler, feature-based models, like regression models, decision trees or Bayesian networks \citep{rudin2019stop, CML_for_healthcare, explainable_trees}. However, these models require large structured tabular datasets as a training resource to ensure their generalization to broader patient populations. As such, developing methods to extend the tabular portion of the EHR with as much information as possible could benefit a wide range of downstream applications. 

In this work, we propose a method for patient-level information extraction that leverages both the structured tabular features available in a patient's EHR and the unstructured clinical notes documenting physician encounters. We model the tabular portion of the EHR using a Bayesian network (BN) with an expert-defined structure and learned probabilities, enabling interpretable integration of background information. This network is then connected to the predictions of a text classifier that interprets the clinical notes. By linking these two modalities, the BN can adjust the text classifier's predictions when they conflict with evidence from the tabular data, and infer missing information when the text is incomplete, all in a manner that remains transparent to the end user.

As a proof-of-concept for our method, we focus on a specific use-case built on the SimSUM dataset \citep{SimSUM}. SimSUM is a simulated benchmark of 10,000 artificial patient records, linking tabular variables (like symptoms, diagnoses and underlying conditions) to associated clinical notes describing the patient encounter in the domain of respiratory diseases. To our knowledge, SimSUM is the only available clinical dataset that provides both tabular and textual data connected through a shared and fully known data-generating process. This inherent alignment between modalities, together with the known BN structure underlying the data, enables fundamental research on integrating tabular and text-based models.

Figure \ref{fig:consistency_overview} shows the overview of our setup and proposed method. On one side, we have a tabular EHR containing encoded background, diagnoses, and treatment. In addition, we have a clinical text describing the symptoms experienced by this patient. Since such symptom information can be valuable for downstream applications (e.g., automated diagnostic systems), our goal is to incorporate it into the EHR in a structured format. In a standard information extraction pipeline, a neural text classifier -- however advanced -- is typically used to predict whether a symptom is mentioned in the text. Our approach additionally exploits the tabular information already present in the EHR by connecting it to the target concepts for extraction (i.e.,~symptoms) through a BN. The relations in the BN (i.e.~which underlying conditions may give rise to which symptoms) are provided by an expert, while the exact probabilities are learned from the tabular portion of the EHR. 

Our main contribution lies in \textbf{enabling patient-level information extraction}, by \textbf{combining the predictions made by a Bayesian network and a text classifier in a probabilistic manner}, augmenting the established approach of virtual evidence with an additional \textit{consistency node}. 
This combination enables our method to correct for inconsistencies and missing information in the text by leveraging tabular data and background knowledge through the BN. The consistency node improves the calibration of the final predictions compared to using virtual evidence alone, especially in abnormal cases where information is missing from the text, while retaining high predictive power in common cases.
Moreover, by obtaining a probabilistic rather than deterministic encoding of the symptoms within the tabular record, our approach yields extracted information that is more robust for downstream use. Consequently, users of the final tabular dataset can either apply a threshold to obtain hard labels for these concepts or directly utilize the probabilistic outputs, depending on their specific application and system requirements.

The remainder of our work is structured as follows. In Section \ref{sec:related_work} we discuss related work, including an in-depth comparison of our work with previous approaches for integrating BNs with neural text classifiers. Then, Section \ref{sec:methods} explains the building blocks of our multimodal approach and how these are connected together using the consistency node. In Section \ref{sec:results}, we report the performance of our method on the SimSUM dataset for various sample sizes, and take a closer look at how our multimodal approach improves over uni-modal baselines. Finally, Sections \ref{sec:conclusion} and \ref{sec:limitations} summarize the potential of our method, while also addressing its limitations. The code for this project can be found at \url{https://github.com/AdrickTench/patient-level-IE}.


\section{Related work} \label{sec:related_work}

In Section \ref{sec:multi_modal_EHRs}, we first introduce related work on representation learning for electronic health records (EHRs), in particular focusing on multimodal approaches that integrate tabular data and text, as well as those that incorporate background knowledge. Section \ref{sec:integrating_BNs_text} then zooms in on two closely related methods that integrate text in BNs, contrasting them with our approach. Finally, Section \ref{sec:virtual_evidence} explains the concept of virtual evidence.

\subsection{Multimodal EHR Representation Learning} \label{sec:multi_modal_EHRs} 

\paragraph{Modeling tabular data and text}  Many studies focus on combining two modalities commonly found in EHRs --- structured tabular data (e.g. disease codes, patient demographics, treatment codes, lab results) and unstructured text (e.g. discharge summaries or nursing notes) -- for representation learning. Most methods use state-of-the-art encoders to learn a representation for each modality, combining both through simple concatenation \citep{combining_structured_unstructured}, ensemble methods \citep{multimodal_ICD} or an attention mechanism \citep{liu2022multimodal, carer, hierarchical_multimodal}. 
While our work also focuses on combining these two modalities, our goal is not to learn a black-box patient-level representation for downstream prediction tasks. Instead, we aim to enrich the structured portion of the EHR with information extracted from the text, thereby facilitating its use in interpretable downstream prediction models such as regression models, decision trees, or BNs \citep{rudin2019stop}. Furthermore, by providing a probabilistic view on the extracted structured variables, our approach enables intermediate manual inspection of the dataset before it is used in downstream models, which is not possible with black-box patient representations.
\paragraph{Integrating background knowledge} 
Particularly interesting to our work are representation learning methods that include some form of background knowledge to improve the learned multimodal representations. \citet{carer} enhance their multimodal encoding for the structured and text portions of the EHR with a clinical reasoning embedding, which is obtained by retrieving relevant clinical documents and asking an LLM to reason over them. 
\citet{ramEHR} use a similar information retrieval module to enhance disease code embeddings.

Instead of incorporating relevant information from clinical corpora, we focus on incorporating graph-based background knowledge. In this line of work, both \citet{mime} and \citet{graph_text_multimodal} model the structured portion of the EHR with a graph attention network, where each admission is represented as a hierarchical graph containing procedure, diagnoses and prescription nodes. As an extension, \citet{learning_graphical_structure} learn to automatically construct the causal structure of a patient admission (in particular, modeling which diagnoses lead to the prescription of which treatment), by training a graph convolutional transformer jointly with supervised medical prediction tasks. This approach is conceptually similar to the BN component of our method. However, in our case, we (i) learn a BN rather than a graph convolutional transformer, and (ii) 
rely on domain experts to specify the medical relationships symbolically, learning only the underlying probability distributions from the data rather than the relations among variables.


Other works incorporate graph-based background knowledge into EHR representation learning by leveraging medical ontologies such as ICD-10, SNOMED-CT, or UMLS, which encode hierarchical clinical concepts and relationships among medical entities \citep{gram}. Most approaches employ graph neural networks to model the relationships between medical codes, integrating these representations into the structured EHR embedding at various stages of the representation learning pipeline \citep{liu2022multimodal, kame, medpath}. In contrast, we do not rely on knowledge graphs to model relationships between medical variables. Instead, we represent background knowledge through a BN, which not only captures clinical dependencies among variables (as knowledge graphs do) but also enables probabilistic reasoning and prediction.

\begin{figure*}[ht!]
  \includegraphics[width=\textwidth]{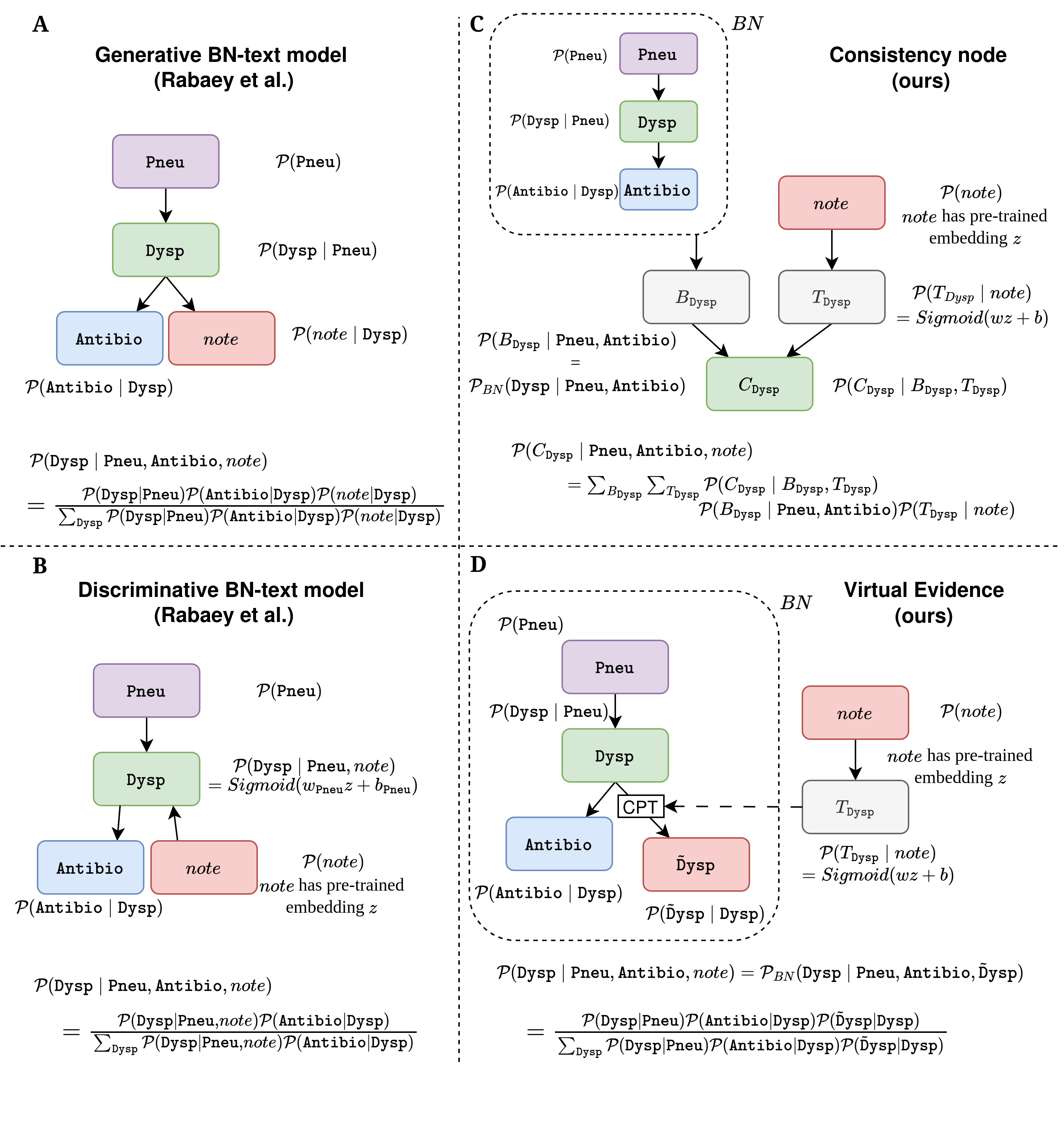}
  \caption{Comparison of our work with \citet{AIME_paloma}. In this running example, we aim to predict the probability $\mathcal{P}(\texttt{Dysp} \mid \texttt{Pneu}, note)$ that a patient suffers from the symptom dyspnea (\texttt{Dysp}), given both tabular information (whether the patient has pneumonia, or \texttt{Pneu}, and was prescribed antibiotics, or \texttt{Antibio}) and a clinical note ($note$). In the generative and discriminative BN-text models proposed by \citet{AIME_paloma} (\textbf{A} and \textbf{B}), a $note$ node is directly integrated in the BN, allowing one to perform Bayesian inference with mixed textual and tabular evidence. In our method, we split off the BN -- where Bayesian inference is performed only over tabular evidence -- and the neural text classifier, integrating their predictions through the consistency node $C_{\texttt{Dysp}}$ (\textbf{C}) and virtual evidence (\textbf{D}). Our method improves on the poor performance of \textbf{A} and poor scalability, interpretability, and causal structure of \textbf{B}. }  
  \label{fig:related_work}
\end{figure*}

\subsection{Integrating Bayesian networks and text} \label{sec:integrating_BNs_text}

\paragraph{Bayesian networks} A BN has the ability to model complex problems involving uncertainty, while combining data with expert knowledge in an interpretable graphical structure \citep{comprehensive_scoping_BNs}. It is made up of two parts: a Directed Acyclic Graph (DAG) modeling the relations between the variables, and a joint probability distribution factorizing into a set of conditional distributions, one for each variable. BNs have proven useful to model a wide range of medical conditions in clinical research settings \citep{BN_healthcare_condition, BN_risk_prediction}, including respiratory diseases such as pneumonia and Covid-19 \citep{BN_respiratory}. While one could automatically learn the structure of the BN from data \citep{learning_graphical_structure, causal_disc_medical}, a particular asset of the BN is the possibility to integrate expert knowledge in the prediction process. In our case, we indeed assume that the relations between the clinical variables have been provided by an expert, while we learn the conditional probabilities from the data. Despite their potential, the clinical adoption of BNs remains limited, largely because they struggle to handle realistic medical data, where unstructured text is abundant \citep{BNs_healthcare_adoption}. Consequently, there is significant potential in developing methods to effectively integrate textual information into BNs. Despite this, there has been limited prior work on integrating text data into BNs. We now zoom in on two highly relevant contributions.


\paragraph{DeepProbLog \citep{deepproblog}} 

One contribution comes from the field of Neuro-Symbolic AI, in the form of the DeepProbLog framework \citep{deepproblog}. Here, a probabilistic logic program is extended with neural predicates, whereby a neural network is used to learn a representation of an unstructured concept (in our case, clinical text), which is further treated as a regular predicate in the program. Since BNs are a symbolic method, they can be naturally formulated as probabilistic logic programs. 
However, one limitation of DeepProbLog is that neural predicates can only serve as root nodes within the BN, from which probabilities propagate downward to fully symbolic, categorical variables. In other words, these neural predicates cannot have any symbolic parent variables. 
In our case, the tabular nodes we aim to predict -- namely, the symptoms -- occupy arbitrary positions within the BN and typically have multiple parent variables.

\paragraph{BN-text model \citep{AIME_paloma}}

To tackle this limitation, \citet{AIME_paloma} propose two approaches -- a \textit{generative} and a \textit{discriminative} BN-text model -- which integrate text directly into a clinical BN, allowing the text to be part of the evidence in the Bayesian inference procedure.
Figure \ref{fig:related_work} compares both approaches to our method, making use of a simple example. In this small BN, the disease pneumonia (\texttt{Pneu}) gives rise to the symptom dyspnea (\texttt{Dysp}), with a clinician deciding whether or not to prescribe antibiotics (\texttt{Antibio}) based on the presence of this symptom in the patient. A clinical $note$ may describe information about the symptom dyspnea. The first approach proposed by \citet{AIME_paloma}, called the generative BN-text model (Figure \ref{fig:related_work}A), directly includes a text node in the BN and models its conditional distribution $\mathcal{P}(note \mid \texttt{Dysp})$ by learning the parameters of a multivariate Gaussian over pre-trained text embeddings. This parametric assumption proved too stringent to work in practice, leading to inferior performance of this method. \citet{AIME_paloma} therefore propose a second approach, called the discriminative BN-text model (Figure \ref{fig:related_work}B). This method deviates from the causal structure of the problem -- the symptom dyspnea influencing the content of the text -- and instead uses a neural classifier to model $\mathcal{P}(\texttt{Dysp} \mid \texttt{Pneu}, note)$. In other words, it extends the DeepProbLog framework \citep{deepproblog} by learning multiple neural predicates, conditional on the possible values of the parent variable \texttt{Pneu}. However, this approach comes with two disadvantages: (i) learning a text classifier for every combination of parent values per symptom does not scale well when symptoms have many parents, and (ii) the effect of the diagnosis \texttt{Pneu} is implicitly encoded in the text classifiers, rather than explicitly injected through the conditional distributions in the BN. 

As illustrated in Figure \ref{fig:related_work}, our solution addresses these limitations by assuming a less stringent connection between the BN modeling the structured tabular variables on the one hand, and the neural text classifier modeling the clinical notes on the other. With \textit{virtual evidence} (Figure \ref{fig:related_work}D, Sections \ref{sec:virtual_evidence} and \ref{sec:symp_virtual_evidence}), a child node is introduced to the BN with a CPT determined by the neural classifier. With the \textit{consistency node} (Figure \ref{fig:related_work}C, Section \ref{sec:consistency_node}), the individual predictions of the BN and neural classifier are integrated by a separate module trained to assign appropriate weights to their contributions. Our final method incorporates both \textit{virtual evidence} and a \textit{consistency node} to combine the predictions of the BN with the neural classifiers (see Figure \ref{fig:consistency_overview} for an overview of the combined model).


\subsection{Virtual evidence}
\label{sec:virtual_evidence}

Virtual evidence is an established method to incorporate the predictions of neural networks with a BN \citep{bnve, hmmve}, and can be understood as evidence with uncertainty \citep{bnsnowexample}. \citet{pearl88} first introduced virtual evidence as a convenient way to incorporate uncertain evidence into a BN. Pearl's method treats virtual evidence as likelihood information. As opposed to the standard hard presence or absence of a piece of evidence in a regular BN, a piece of virtual evidence is represented by a real number in $[0, 1]$ that indicates the confidence in observing a particular value for a variable $V$.\footnote{Note that virtual evidence is sometimes referred to as soft evidence. We avoid using the term soft evidence here, as it also refers to a different type of uncertain evidence \citep{bnsnowexample}.}

To incorporate virtual evidence into a BN, a child node is added to the variable for which virtual evidence is provided. The child node itself is ``observed'' -- i.e., provided as standard, hard evidence -- while its CPT is determined by the uncertain evidence. For example, suppose a binary variable $A$ is observed with a probability of $0.8$. To include this uncertain observation as virtual evidence, a child node $\tilde{A}$ is added to the BN with a CPT encoding $P(\tilde{A}|A)$:\footnote{\label{shaded_BN_footnote}Here we use the convention of shading in the observed nodes in the BN.}

\newsavebox{\tablebox}

\begin{table}[ht]
    \centering
    \sbox{\tablebox}{%
        \begin{minipage}[c]{0.48\linewidth}
            \centering
            \small
            \begin{tabular}{|c|c|c|}
                \hline
                & $A$ & $\neg A$ \\ \hline
                $\tilde{A}\rule{0pt}{1.1em}$ & 0.8 & 0.2 \\ \hline
            \end{tabular}
        \end{minipage}%
    }
    \usebox{\tablebox}%
    \hspace{0.01\linewidth}
    \begin{minipage}[c]{0.48\linewidth}
        \centering
        \includegraphics[height=2.4\ht\tablebox]{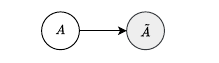}
    \end{minipage}
\end{table}

This method of incorporating uncertain evidence retains the prior probability distribution of the BN. It is possible to provide virtual evidence for multiple variables, or even multiple pieces of virtual evidence for the same variable \citep{bnuncertainupdate}. Furthermore, it is still sensible to query the BN for the variable(s) for which virtual evidence is provided.

To make this concrete, consider a slight extension of our example above. Now $A$ has a single parent $B$, with the following CPT:\footnote{See footnote \ref{shaded_BN_footnote}.}

\newsavebox{\tableboxtwo}

\begin{table}[ht]
    \centering
    \sbox{\tableboxtwo}{%
        \begin{minipage}[c]{0.48\linewidth}
            \centering
            \small
            \begin{tabular}{|c|c|c|}
                \hline
                & $B$ & $\neg B$ \\ \hline
                $A$ & 0.3 & 0.7 \\ \hline
            \end{tabular}
        \end{minipage}%
    }
    \usebox{\tableboxtwo}%
    \hspace{0.01\linewidth}
    \begin{minipage}[c]{0.48\linewidth}
        \centering
        \includegraphics[height=2\ht\tableboxtwo]{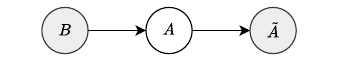}
    \end{minipage}
\end{table}

We can then use standard variable elimination \citep{koller2009probabilistic} to calculate $P(A|B, \tilde{A}) \approx 0.6316$ (Equation \ref{eq:ve_calculation}), where both $B$ and virtual evidence $\tilde{A}$ are provided as evidence to the BN. Note that the final probability is lower than the uncertain evidence of $0.8$ provided through $\tilde{A}$, because it factors in the prior probability $P(A|B)$.

\begin{align}
    P(A \mid B, \tilde{A}) = \frac{P(\tilde{A} \mid A) \, P(A \mid B)}{\sum_{a'} P(\tilde{A} \mid a') \, P(a' \mid B)} = \frac{0.8\cdot0.3}{(0.8\cdot0.3)+(0.2\cdot0.7)} \approx 0.6316
    \label{eq:ve_calculation}
\end{align}

As shown in Figure \ref{fig:consistency_overview}, our method uses virtual evidence but extends it with a \emph{consistency node} (Section \ref{sec:consistency_node}), further improving the calibration of the full model.

\section{Methods} \label{sec:methods}

Following the introduction of the dataset and our experimental setup in Section~\ref{sec:prelim}, we present the fundamental components of our model: a Bayesian network (Section~\ref{sec:BN}) and a neural network (Section~\ref{sec:NN}). The core contribution of this work is the investigation of two strategies for integrating these models' predictions: virtual evidence (Section~\ref{sec:symp_virtual_evidence}) and a consistency node (Section~\ref{sec:consistency_node}).

\subsection{Preliminaries} \label{sec:prelim}

\paragraph{Dataset} The SimSUM dataset links artificial tabular patient records with artificial clinical notes describing a patient's visit to the general practitioner's office. By design, the clinical concepts expressed in the text (the symptoms experienced by the patient) and the tabular background information are connected through a BN representing domain knowledge. As shown in Figure \ref{fig:consistency_overview}, the BN relates two respiratory diseases (\texttt{Pneumonia} and \texttt{Common cold}) with their associated symptoms (\texttt{Dyspnea}, \texttt{Cough}, \texttt{Pain}, \texttt{Nasal} symptoms and \texttt{Fever}), as well as including underlying respiratory conditions (\texttt{Asthma}, \texttt{Smoking}, \texttt{COPD} and \texttt{Hay fever}), some background (\texttt{Season} of the year), and a treatment and outcome variable (whether \texttt{Antibiotics} were prescribed and the \texttt{\#Days} the patient ended up staying at home as a result of their symptoms).\footnote{In the original SimSUM dataset, two other variables are included (\texttt{policy} and \texttt{self-employed}). We leave these out of our setup as they are non-clinical variables that would not be encoded in the patient record in a realistic setting.} All variables are binary, except for \texttt{Fever}, which has three levels (none, low and high) and \texttt{\#Days}, which is discrete and bounded ($0,\dots,15$). 

The SimSUM notes are generated in such a way that they describe the five symptoms and occasionally mention underlying respiratory conditions, but they do not explicitly mention the diagnoses. In this way, both modalities (tabular features and clinical notes) contain complementary information. Every patient has one unique note, with two versions: ``normal'' and ``compact''.
We will report results on the normal notes. 

\paragraph{Setup} As shown in Equation \ref{eq:setup}, we have a set $\mathcal{X}$ of $n$ patient records where each patient $X^{(i)}$ is described by their tabular record $tab^{(i)}$, their clinical note $note^{(i)}$ and their set of symptoms $s^{(i)}$.

\begin{small}
\begin{align}
    \mathcal{X} = &\{X^{(0)}, ..., X^{(n-1)} \mid X^{(i)} = \{tab^{(i)}, note^{(i)}, s^{(i)}\}\} \nonumber\\
    tab^{(i)} = &\{\texttt{Asthma} = t_{0}^{(i)}, \texttt{Smoking} = t_{1}^{(i)}, \texttt{COPD} = t_{2}^{(i)}, \;
    \texttt{Hay\;fever} = t_{3}^{(i)}, \texttt{Season} = t_{4}^{(i)}, \nonumber\\
    &\texttt{Pneumonia} = t_{5}^{(i)}, \texttt{Common\;cold} = t_{6}^{(i)}, \texttt{Antibiotics} = t_{7}^{(i)}, \texttt{\#Days} = t_{8}^{(i)}\} \nonumber\\
    s^{(i)} = &\{\texttt{Dyspnea} = s_{0}^{(i)}, \texttt{Cough} = s_{1}^{(i)}, \texttt{Pain} = s_{2}^{(i)}, \;
    \texttt{Nasal} = s_{3}^{(i)}, \texttt{Fever} = s_{4}^{(i)}\} \nonumber\\
    = &\{s_{j} = s_{j}^{(i)} \mid j = 0,\dots,4\}
\label{eq:setup}
\end{align}
\end{small}
Here, $s_{j}^{(i)} \in \mathcal{V}_{s_j}$, where $\mathcal{V}_{s_j} = \{yes, no\}$ for $j = 0\dots 3$ (symptoms \texttt{Dyspnea}, \texttt{Cough}, \texttt{Pain} and \texttt{Nasal}), and $\mathcal{V}_{s_j} = \{high, low, none\}$ for $j=4$ (symptom \texttt{Fever}).

While the tabular values $t_{0}^{(i)},\dotsc,t_{8}^{(i)}$ and the text $note^{(i)}$ are available in the patient record at all times, the symptom values $s_{0}^{(i)},\dotsc,s_{4}^{(i)}$ are only available during training. Out of the full set of patient records $\mathcal{X}$, we select a training set $\mathcal{X}_{train}$, 
and a test set $\mathcal{X}_{test}$. At test time, we aim to predict the probability for each of the symptoms given the tabular data and text note, i.e., the distributions $\mathcal{P}(s_{j} \mid tab^{(i)}, note^{(i)})$ for $X^{(i)} \in \mathcal{X}_{test}$ and $j = 0\dots 4$. 

Note that the distinction between the symptoms $s^{(i)}$ and tabular data $tab^{(i)}$ serves merely to denote $s^{(i)}$ as the targets for information extraction. While targeting $s^{(i)}$ is a natural choice given the data-generating process behind the SimSUM dataset, our method does not depend on any specific choice of target variables for extraction.

\subsection{Bayesian network} \label{sec:BN}

We model the tabular portion of the data with a BN, where the relations between the variables (the DAG, as shown in Figure \ref{fig:consistency_overview}) are provided up-front by an expert. This DAG naturally prescribes how the joint probability distribution factorizes into conditional probability distributions (CPD), one for each variable conditional on its parents. As in \citet{SimSUM}, each CPD is learned independently by fitting it to the tabular portion of the training data $\mathcal{X}_{train}$. We refer to Appendix \ref{sec:app_training_BN} for more details on this training procedure. 

When the full joint distribution defined by the BN has been learned, we can perform Bayesian inference to obtain $\mathcal{P}(s_{j} \mid tab^{(i)})$ for each patient $i \in \mathcal{X}_{test}$ and each symptom $j$, by filling in the tabular evidence available for this patient and performing variable elimination \citep{koller2009probabilistic}. For example, for the patient in Figure \ref{fig:consistency_overview}, we would calculate $\mathcal{P}(\text{\texttt{Dyspnea} = yes} \mid \text{\texttt{Asthma} = yes}, \dots, \text{\texttt{\#Days} = 5})$ 
by summing out the other symptoms \texttt{Cough}, \texttt{Pain}, \texttt{Nasal}, and \texttt{Fever} (which are unavailable at test time) from the joint distribution 
and normalizing. From now on, we will denote the BN's prediction for the symptom $s_j$ as $\mathcal{P}(B_{s_j} \mid tab^{(i)})$.

\subsection{Neural network} \label{sec:NN}

To model the text portion of the data, we use a lightweight neural text classifier. For simplicity, we use the same encoder and architecture as SimSUM (\citep{SimSUM}), described below.\footnote{Note that we only intend to provide an example of a lightweight text classifier for comparison and do not require the best possible model.} We train separate classifiers for each symptom. At the input, the clinical note is first split into sentences. Each sentence is transformed into an embedding using the pretrained clinical representation model BioLORD-2023 \citep{biolord}, after which all sentence embeddings are averaged to obtain a single representation for the full note. This note embedding is then fed into a multi-layer perceptron with one hidden layer, of which the weights are trained using the cross-entropy objective over the symptom labels in $\mathcal{X}_{train}$. We finally obtain class probabilities by applying a Sigmoid activation (or Softmax for \texttt{Fever}, which has three classes) to the output layer. At test-time, each trained classifier can provide predictions $\mathcal{P}(s_{j} \mid note^{(i)})$ for symptom $j$, for each patient $i \in \mathcal{X}_{test}$. From now on, we will denote the text classifier's prediction for the symptom $s_j$ as $\mathcal{P}(T_{s_j} \mid note^{(i)})$.


\subsection{Virtual evidence}\label{sec:symp_virtual_evidence}

To combine the probabilities of the BN and neural classifiers, we can provide the latter's predictions to the BN as virtual evidence. That is, for a patient $i \in \mathcal{X}$, the predictions of the neural classifiers for each symptom $\mathcal{P}(T_{s_j} \mid note^{(i)})$ are provided to the BN as virtual evidence, as outlined in Section \ref{sec:virtual_evidence}. We refer to the virtual evidence for a symptom \texttt{s} with $\tilde{\texttt{s}}$. For example, $\tilde{\texttt{D}}\texttt{yspnea}^{(i)}$ refers to the virtual evidence for the presence of \texttt{Dyspnea} in patient $i$. The combined tabular and text prediction for each symptom can be noted as follows, where both the tabular evidence $tab^{(i)}$ and the virtual evidence for all symptoms are provided:
\begin{small}
\begin{align}
    \mathcal{P}(s_j \mid tab^{(i)}, \tilde{\texttt{D}}\texttt{yspnea}^{(i)}, \tilde{\texttt{C}}\texttt{ough}^{(i)}, \;
    \tilde{\texttt{P}}\texttt{ain}^{(i)}, \tilde{\texttt{N}}\texttt{asal}^{(i)}, \tilde{\texttt{F}}\texttt{ever}^{(i)}) \nonumber
\end{align}
\end{small}
This probability can be obtained through variable elimination, again by summing out the other (non-virtual) symptom nodes, apart from target symptom $s_j$. From now on, we will denote the prediction of the BN with virtual evidence for the symptom $s_j$ as $\mathcal{P}(V_{s_j} \mid tab^{(i)}, note^{(i)})$.

\subsection{Consistency node} \label{sec:consistency_node}

After training both the BN and the neural text classifier over $\mathcal{X}_{train}$, we can obtain $\mathcal{P}(B_{s_j} \mid tab^{(i)})$ (prediction based on tabular evidence) and $\mathcal{P}(T_{s_j} \mid note^{(i)})$ (prediction based on text evidence), for any patient $i \in \mathcal{X}_{test}$ and any symptom $j$. From now on, we will leave out the patient index $i$ to avoid cluttering the notation. We now combine both probabilities through a consistency node $C_{s_j}$, as illustrated in the overview in Figure \ref{fig:consistency_overview}. The nodes $B_{s_j}$, $T_{s_j}$ and $C_{s_j}$ form a probabilistic graphical model, with joint distribution as shown in Equation \ref{eq:consistency_joint}.


\begin{small}
\begin{align}
    \mathcal{P}(C_{s_j}, B_{s_j}, T_{s_j} \mid tab, note) 
    &= \mathcal{P}(B_{s_j} \mid tab)\;
       \mathcal{P}(T_{s_j} \mid note)\;
       \mathcal{P}(C_{s_j} \mid B_{s_j}, T_{s_j})
    \label{eq:consistency_joint}
\end{align}
\end{small}

To obtain $\mathcal{P}(C_{s_j} \mid tab, note)$, we simply need to marginalize out the BN and text classifier predictions from the joint distribution, as in Equation \ref{eq:consistency_sum}. We do this by summing over all possible values of the symptom ${s_j}$: 


\begin{footnotesize}
\begin{align}
    &\mathcal{P}(C_{s_j} \mid tab, note)
    =\sum_{b'}\sum_{t'}\left[\mathcal{P}(B_{s_j} = b' \mid tab)\mathcal{P}(T_{s_j} = t' \mid note) 
    \mathcal{P}(C_{s_j} \mid b', t')\right]\text{, with} \; b', t' \in V_{s_j}
\label{eq:consistency_sum}
\end{align}
\end{footnotesize}

The conditional distribution $\mathcal{P}(C_{s_j} \mid B_{s_j}, T_{s_j})$ can be computed over the training set $\mathcal{X}_{train}$.\footnote{To prevent overfitting on the training set, which would taint the conditional distribution, we train the neural networks using 5-fold cross-validation as described in Appendix \ref{app:neural_text}.} We first use the BN and neural classifier to obtain the probabilities $\mathcal{P}(B_{s_j} =b'\mid tab^{(k)})$ and $\mathcal{P}(T_{s_j} =t'\mid note^{(k)})$ for each patient $k \in \mathcal{X}_{train}$, each symptom $j$ and each label $b', t' \in V_{s_j}$. As shown in Equations \ref{eq:consistency_weight_fractions} and \ref{eq:consistency_weight}, we then calculate the agreement of the predicted probabilities with respect to the ground truth label $c'$ as observed in $\mathcal{X}_{train}$:

\begin{footnotesize}
\begin{align}
    \mathcal{P}(C_{s_j}=c'\mid b', t') \;
    &= \frac{W\{c', b', t'\}}{\sum_{c''}W\{c'', b', t'\}} \text{, with $c', c'' \in V_{s_j}$}\label{eq:consistency_weight_fractions}
\end{align}
\end{footnotesize}
where the weights are obtained as:
\begin{footnotesize}
\begin{align}
    W\{c', b', t'\} \;
    &= \sum_{k \in \mathcal{X}_{train}} \mathcal{P}(B_{s_j} =b' \mid tab^{(k)}) \;
    \cdot\mathcal{P}(T_{s_j} =t' \mid note^{(k)})\;
    \cdot\ind\big[s_j^{(k)} = c'\big]
\label{eq:consistency_weight}
\end{align}
\end{footnotesize}

Intuitively, this process calculates the agreement between the BN and the text classifier on the training set $\mathcal{X}_{train}$. When they disagree (meaning $b' \neq t'$), one of the two models will be right more often. For example, if the text classifier agrees with the ground truth label in the training set more often, this means $\mathcal{P}(c' = t' \mid b', t') > \mathcal{P}(c' = b' \mid b', t')$. The terms containing the factor $\mathcal{P}(T_{s_j}=t' \mid note)$ will then receive a higher weight in Equation \ref{eq:consistency_sum}, pushing the prediction towards the label $t'$. We provide a simple numerical example in Appendix \ref{sec:numerical_example}. 

From Section \ref{sec:symp_virtual_evidence} we can also obtain $\mathcal{P}(V_{s_j} \mid tab, note)$, denoting the prediction of the BN with virtual evidence for the symptom $s_j$. In the discussion above, $\mathcal{P}(V_{s_j} \mid tab, note)$ can be used in place of $\mathcal{P}(B_{s_j} \mid tab)$, in particular in Equation \ref{eq:consistency_joint}. This leads to our final model shown in Figure \ref{fig:consistency_overview}, where we calculate the consistency between the virtual evidence-enhanced BN and the text classifier prediction, also called \textbf{V-C-BN-text} in the next section.

\section{Empirical Results} \label{sec:results}

We split the SimSUM dataset into a training set $\mathcal{X}_{train}$ of 8000 samples 
and a test set $\mathcal{X}_{test}$ of 2000 samples. To investigate the performance of our method in various training regimes, we subsample different training sets $\mathcal{X}_{train}^{n}$ from $\mathcal{X}_{train}$, logarithmically sized between $n=100$ and $8000$, i.e. 
$n \in \{100, 187, 350, 654, 1223, 2287, 4278, 8000\}$, using $20$ different seeds per $n$. We also use these seeds for initialization of the model parameters and weights. For more details on hyperparameter tuning, we refer to Appendix \ref{sec:app_training}. The BN and neural text classifier are trained on the tabular and text portion of $\mathcal{X}_{train}^{n}$, respectively, across these $20$ seeds. We compare the following model architectures:
\begin{itemize}
    \setlength{\itemsep}{1pt}
    \setlength{\parskip}{0pt}
    \setlength{\parsep}{0pt}
    \item \textbf{BN-only}: The BN predicts $\mathcal{P}(B_{s_j} \mid tab^{(i)})$ for every patient $i$, taking tabular evidence $tab^{(i)}$ as input (Section \ref{sec:BN}).
    \item \textbf{text-only}: The neural text classifier predicts $\mathcal{P}(T_{s_j} \mid note^{(i)})$ for every patient $i$, taking text evidence $note^{(i)}$ as input (Section \ref{sec:NN}).
    \item \textbf{V-BN-text}: The \textbf{BN-only} and \textbf{text-only} predictions are combined by using the text-only predictions as virtual evidence, obtaining the combined prediction $\mathcal{P}(V_{s_j} \mid tab^{(i)}, note^{(i)})$ (Section \ref{sec:symp_virtual_evidence}).
    \item \textbf{C-BN-text}: We learn the distribution of the consistency node on $\mathcal{X}_{train}^{n}$ for the \textbf{BN-only} and \textbf{text-only} predictions, obtaining a final consistent prediction that symptom $s_j$ is present in patient $i$: $\mathcal{P}(C_{s_j} \mid tab^{(i)}, note^{(i)})$ (Section \ref{sec:consistency_node}).
    \item \textbf{V-C-BN-text}: This approach uses both virtual evidence and the consistency node, corresponding to our final model in Figure \ref{fig:consistency_overview}. We learn the distribution of the consistency node on $\mathcal{X}_{train}^{n}$ for the \textbf{V-BN-text} and \textbf{text-only} predictions, obtaining the combined prediction $\mathcal{P}(VC_{s_j} \mid tab^{(i)}, note^{(i)})$ (Section \ref{sec:symp_virtual_evidence} and \ref{sec:consistency_node}). Note the contrast with \textbf{C-BN-text}, which uses $\mathcal{P}(B_{s_j} \mid tab^{(i)})$ rather than $\mathcal{P}(V_{s_j} \mid tab^{(i)}, note^{(i)})$.
\end{itemize}

The models \textbf{C-BN-text} and \textbf{V-BN-text}, can be viewed as ablations of the final model \textbf{V-C-BN-text}, using only one of the consistency node or virtual evidence to combine the neural classifiers with the BN. 

In addition to these ablations, we consider the following additional baseline:

\begin{itemize}
    \setlength{\itemsep}{1pt}
    \setlength{\parskip}{0pt}
    \setlength{\parsep}{0pt}
    \item \textbf{Concat-tab-text}: This early-fusion baseline concatenates the tabular features and the text embedding at the input of an MLP with the same architecture as the \textbf{text-only} classifier. We use the same implementation as described in \citet{SimSUM}, where binary variables are transformed into a one-hot encoding, and the variable \texttt{\#Days} is preprocessed using standard scaling. In contrast to the 
    other fusion models we explore, this black-box model treats the tabular and text features as a single entity, and does not involve a BN. For more details on the implementation of this baseline, we refer to Appendix \ref{app:multi_modal_base}.
\end{itemize}

To evaluate these models, we calculate the \textbf{average precision} and \textbf{Brier score} over $\mathcal{X}_{test}$, across $20$ seeds. We report average precision (equivalent to area under the precision-recall curve), rather than some threshold-based metrics like F1 score or accuracy, because our goal is to provide a probabilistic estimate of the symptom $s_j$ being present in the patient, rather than a hard decision. We choose average precision over area under the ROC curve since the former is better equipped to deal with imbalanced datasets \citep{precision_recall_ROC}, which is indeed the case for the symptom labels in SimSUM. Since our classifiers for \texttt{Fever} have three classes, we report the macro average precision in that case. 

We also report the Brier score \citep{brier} (equivalent to the MSE between the predicted probabilities and ground truth labels for binary symptoms) to reflect the accuracy of our models' predicted probabilities. It provides a combined measure of calibration and confidence. Like average precision, this metric quantifies the correctness of the predicted probabilities without requiring a hard threshold be set for classification. Note that while our Brier scores for the binary symptoms lie in the $[0,1]$ range, we did not scale the score for \texttt{Fever} (ternary), so it remains in the $[0, 2]$ range.

Section \ref{sec:overall_results} presents our main results, where we compare our multimodal consistency method with the uni-modal and multimodal baselines. Then, Section \ref{sec:pvm_subsets} unravels where the improvements lie by examining the models' performance on distinct subsets of the test set. Following this, we investigate how our consistency method handles a shift in text data distribution during inference in Section \ref{sec:data_shift}. Our key takeaways are summarized in Section \ref{sec:takeaways}.

\subsection{Overall model comparison} \label{sec:overall_results}

As shown in Table \ref{tab:neural_classifiers_baseline}, the naive \textbf{text-only} method already performs quite well for the symptoms \texttt{Dysp}, \texttt{Cough}, and \texttt{Nasal}. Therefore, we focus our analysis on the more difficult symptoms \texttt{Pain} and \texttt{Fever} that show more room for improvement. Tables \ref{tab:comparison_models_average_precision} and \ref{tab:comparison_models_Brier}, respectively, report the average precision and Brier scores for these symptoms. Detailed results for the other symptoms are available in Appendix \ref{sec:app_all_symptom_results}.

\begin{table*}[t]
  \caption{Average precision ($\upuparrows$) for the predictions of the \textbf{text-only} model over $\mathcal{X}_{test}$, averaged over 20 seeds for various training sizes $n$. The symptoms \texttt{Pain} and \texttt{Fever} show the most room for improvement.}
  \label{tab:neural_classifiers_baseline}
\resizebox{\textwidth}{!}{
\begin{tabular}{llcccccccc}
    \toprule
    & & \multicolumn{8}{c}{Training size $n$} \\ \cmidrule{3-10}
    & & \textbf{100} & \textbf{187} & \textbf{350} & \textbf{654} & \textbf{1223} & \textbf{2287} & \textbf{4278} & \textbf{8000}\\
    \midrule
    \texttt{dysp} & & $92.46\pm1.88$ & $94.82\pm1.14$ & $95.78\pm0.49$ & $96.7\pm0.3$ & $97.31\pm0.28$ & $97.98\pm0.16$ & $98.31\pm0.13$ & $98.78\pm0.12$ \\
    \midrule
    \texttt{cough} & & $90.2\pm3.11$ & $94.52\pm0.65$ & $95.71\pm0.48$ & $96.89\pm0.38$ & $97.8\pm0.23$ & $98.26\pm0.14$ & $98.63\pm0.11$ & $98.9\pm0.08$ \\
    \midrule
    \texttt{pain} & & $61.81\pm9.03$ & $72.52\pm4.21$ & $76.48\pm2.61$ & $80.4\pm0.95$ & $82.5\pm0.85$ & $83.77\pm0.88$ & $84.71\pm0.89$ & $86.08\pm0.24$ \\
    \midrule
    \texttt{nasal} & & $95.12\pm0.59$ & $95.84\pm0.56$ & $96.57\pm0.25$ & $97.06\pm0.17$ & $97.43\pm0.18$ & $97.79\pm0.11$ & $98.0\pm0.1$ & $98.16\pm0.03$ \\
    \midrule
    \texttt{fever} & & $69.05\pm4.81$ & $75.01\pm4.12$ & $79.73\pm2.34$ & $86.46\pm0.98$ & $89.86\pm0.65$ & $91.64\pm0.51$ & $93.15\pm0.25$ & $93.93\pm0.21$ \\
    \bottomrule
    \bottomrule
\end{tabular}}
\end{table*}

\begin{table*}[t]
  \caption{Average precision ($\upuparrows$) for the predictions of our models over $\mathcal{X}_{test}$, averaged over 20 seeds for various training sizes $n$. The best model per training size and per symptom is highlighted in \textbf{bold}. The best baseline model for each class is \underline{underlined}. Cases where a model outperforms the best baseline model significantly are indicated by \textbf{*} ($p < 0.05$ in a one-sided Wilcoxon signed-rank test over 20 seeds). \textbf{Change vs. baseline} compares the difference between the strongest baseline and non-baseline models.}
  \label{tab:comparison_models_average_precision}
\resizebox{\textwidth}{!}{
\begin{tabular}{llcccccccc}
    \toprule
    & & \multicolumn{8}{c}{Training size $n$} \\ \cmidrule{3-10}
    & & \textbf{100} & \textbf{187} & \textbf{350} & \textbf{654} & \textbf{1223} & \textbf{2287} & \textbf{4278} & \textbf{8000}\\
    \midrule
    \texttt{pain} & \textbf{BN-only} & 0.3197 & 0.3277 & 0.3409 & 0.3464 & 0.35 & 0.3509 & 0.3517 & 0.3515 \\
    & \textbf{text-only} & \textbf{\underline{0.6181}} & \underline{0.7252} & \underline{0.7648} & \underline{0.804} & \underline{0.825} & 0.8377 & 0.8471 & 0.8608 \\
    & \textbf{Concat-text-tab} & 0.5091 & 0.6537 & 0.7422 & 0.7858 & 0.8183 & \underline{0.8412} & \underline{0.8559} & \underline{0.868} \\
    \cline{2-10}
    & \textbf{C-BN-text} & 0.6135 & \textbf{*0.7271} & \textbf{*0.7723} & *0.8124 & *0.8323 & 0.8435 & 0.8521 & 0.8653 \\
    & \textbf{V-BN-text} & 0.5667 & 0.7027 & 0.7532 & \textbf{*0.8194} & \textbf{*0.8463} & \textbf{*0.8598} & \textbf{*0.8699} & \textbf{*0.8826} \\
    & \textbf{V-C-BN-text} & 0.6022 & 0.7244 & 0.7673 & *0.8146 & *0.8375 & *0.8511 & *0.8606 & *0.8738 \\
    \cline{2-10}
    & \textbf{change vs. baseline} & -0.46\% & +0.19\% & +0.76\% & +1.54\% & +2.13\% & +1.86\% & +1.4\% & +1.46\% \\
    \midrule
    \texttt{fever} & \textbf{BN-only} & 0.4792 & 0.4983 & 0.5167 & 0.5309 & 0.5398 & 0.5437 & 0.5465 & 0.5474 \\
    & \textbf{text-only} & \textbf{\underline{0.6905}} & \underline{0.7501} & \underline{0.7973} & \underline{0.8646} & \underline{0.8986} & 0.9164 & 0.9315 & 0.9393 \\
    & \textbf{Concat-text-tab} & 0.6605 & 0.7495 & 0.7939 & 0.8526 & 0.8951 & \underline{0.922} & \underline{0.9381} & \underline{0.9501} \\
    \cline{2-10}
    & \textbf{C-BN-text} & 0.66 & 0.7472 & 0.7999 & *0.8705 & *0.9017 & 0.9199 & 0.9345 & 0.9434 \\
    & \textbf{V-BN-text} & 0.6521 & 0.7545 & *0.8091 & \textbf{*0.8829} & \textbf{*0.9141} & \textbf{*0.931} & \textbf{*0.9475} & \textbf{*0.9562} \\
    & \textbf{V-C-BN-text} & 0.6644 & \textbf{*0.7595} & \textbf{*0.8107} & *0.8802 & *0.9101 & *0.9267 & *0.9421 & *0.9515 \\
    \cline{2-10}
    & \textbf{change vs. baseline} & -2.61\% & +0.94\% & +1.35\% & +1.82\% & +1.54\% & +0.9\% & +0.94\% & +0.61\% \\
    \bottomrule
    \bottomrule
\end{tabular}}
\end{table*}

\begin{table*}[t]
  \caption{Brier scores ($\downdownarrows$) for the predictions of our models over $\mathcal{X}_{test}$, averaged over 20 seeds for various training sizes $n$. The best model per training size and per symptom is highlighted in \textbf{bold}. The best baseline model for each class is \underline{underlined}. Cases where a model outperforms the best baseline model significantly are indicated by \textbf{*} ($p < 0.05$ in a one-sided Wilcoxon signed-rank test over 20 seeds). \textbf{Change vs. baseline} compares the difference between the strongest baseline and non-baseline models.}
  \label{tab:comparison_models_Brier}
\resizebox{\textwidth}{!}{
\begin{tabular}{llcccccccc}
    \toprule
    & & \multicolumn{8}{c}{Training size $n$} \\ \cmidrule{3-10}
    & & \textbf{100} & \textbf{187} & \textbf{350} & \textbf{654} & \textbf{1223} & \textbf{2287} & \textbf{4278} & \textbf{8000}\\
    \midrule
    \texttt{pain} & \textbf{BN-only} & 0.1138 & 0.1119 & 0.1091 & 0.1078 & 0.1074 & 0.1073 & 0.1071 & 0.1072 \\
    & \textbf{text-only} & \textbf{\underline{0.0854}} & \underline{0.0711} & \textbf{\underline{0.0628}} & \underline{0.0553} & \underline{0.0491} & \underline{0.0451} & \underline{0.0426} & \underline{0.0383} \\
    & \textbf{Concat-text-tab} & 0.0942 & 0.0789 & 0.0666 & 0.0623 & 0.0524 & 0.0478 & 0.0436 & 0.0393 \\
    \cline{2-10}
    & \textbf{C-BN-text} & 0.0915 & 0.0734 & 0.0702 & 0.0604 & 0.0535 & 0.0501 & 0.047 & 0.0423 \\
    & \textbf{V-BN-text} & 0.0963 & 0.0792 & 0.0717 & 0.0577 & 0.0511 & 0.046 & 0.0427 & 0.0385 \\
    & \textbf{V-C-BN-text} & 0.0866 & \textbf{*0.0704} & 0.0647 & \textbf{0.0539} & \textbf{0.0483} & \textbf{0.0444} & \textbf{0.0414} & \textbf{0.0377} \\
    \cline{2-10}
    & \textbf{change vs. baseline} & +0.13\% & -0.07\% & +0.19\% & -0.14\% & -0.08\% & -0.06\% & -0.13\% & -0.06\% \\
    \midrule
    \texttt{fever} & \textbf{BN-only} & 0.3253 & 0.3153 & 0.3083 & 0.3049 & 0.3029 & 0.3016 & 0.301 & 0.3008 \\
    & \textbf{text-only} & \underline{0.256} & 0.2257 & 0.201 & 0.1744 & \underline{0.1448} & 0.1256 & 0.1126 & 0.0984 \\
    & \textbf{Concat-text-tab} & 0.2632 & \underline{0.2208} & \underline{0.1978} & \underline{0.1738} & 0.1449 & \underline{0.1243} & \underline{0.1081} & \underline{0.0962} \\
    \cline{2-10}
    & \textbf{C-BN-text} & 0.2683 & 0.2253 & 0.2089 & 0.1709 & 0.1441 & 0.1261 & 0.1144 & 0.1038 \\
    & \textbf{V-BN-text} & 0.2868 & 0.2327 & 0.2098 & *0.1613 & *0.1353 & 0.1242 & 0.1067 & 0.0986 \\
    & \textbf{V-C-BN-text} & \textbf{0.2535} & \textbf{*0.2115} & \textbf{*0.1911} & \textbf{*0.1511} & \textbf{*0.1269} & \textbf{*0.1146} & \textbf{*0.1014} & \textbf{0.0941} \\
    \cline{2-10}
    & \textbf{change vs. baseline} & -0.25\% & -0.93\% & -0.67\% & -2.27\% & -1.79\% & -0.96\% & -0.68\% & -0.22\% \\
    \bottomrule
    \bottomrule
\end{tabular}}
\end{table*}


Examining Tables \ref{tab:comparison_models_average_precision} and \ref{tab:comparison_models_Brier}, we note that the \textbf{BN-only} model performs sub-par to the models involving text, which is unsurprising, as this model has no access to the clinical notes containing ample detail on the presence of the symptoms. More importantly, the results show that the \textbf{V-C-BN-text} model almost always outperforms the \textbf{text-only} and \textbf{Concat-tab-text} models. In terms of Brier scores, the \textbf{V-C-BN-text} model always outperforms its ablated versions \textbf{V-BN-text} and \textbf{C-BN-text}. While the improvements in average precision and Brier score over the baselines are sometimes marginal, they are mostly significant across 20 seeds according to a paired statistical test. As we will show in Section \ref{sec:pvm_subsets}, including knowledge from the tabular portion of the data with help of the BN, allows the \textbf{V-C-BN-text} model to flag mistakes made by the \textbf{text-only} classifier and thereby reliably improve upon its predictions, without impacting its performance on more straightforward cases.


While one might expect lower training sizes to have the most potential for improvement, as the text classifier does not have many examples to learn from in that case, this is not reflected in the results. This is because the BN's performance suffers in these low regimes as well, rendering its predictions less reliable and negatively affecting the performance of the combined models. However, note that we do see the greatest improvements on ``middling'' training sizes (654, 1223, and 2287 samples), where the BN begins to fit the data quite well but the neural classifiers still struggle somewhat. Furthermore, in Appendix \ref{sec:app_GT_BN}, we show that a combined model which has access to the full ground truth BN (including the probabilities, rather than learning them from data), indeed shows larger improvements for smaller training sizes.

Zooming in on the \textbf{Concat-tab-text} baseline, we note that it suffers from overfitting for smaller train sizes, as a result of its naive concatenation of all features. We find that the combined models using the BN are more robust across train sizes and symptoms. These combined models are also much more interpretable, thanks to their modularity, their reliance on the expert-informed BN for the inclusion of the tabular features, and their interpretable late-fusion of the tabular and text predictions using the consistency node and virtual evidence.

\subsection{Analysis of test subsets} \label{sec:pvm_subsets}

To get a better idea of how our combined model \textbf{V-C-BN-text} manages to improve over the \textbf{text-only} baseline, we break the test set up into four distinct subsets based on whether the symptom is present in the patient (treating low + high as present for \texttt{Fever}) and whether the symptom is mentioned in the text: \{\textit{(present, mentioned)}; \textit{(present, not mentioned)}; \textit{(not present, mentioned)}; \textit{(not present, not mentioned)}\}.\footnote{This is possible in SimSUM thanks to the addition of a label that indicates if a given symptom is mentioned in the text of a clinical note; Appendix \ref{sec:mentions_label} outlines the process for obtaining this label.}

As each subset now only contains either negative or positive examples, it is no longer possible to report the average precision. Therefore, we look only at the Brier score. Table \ref{tab:comparison_models_brier_subsets_pain_fever} provides the results for these subsets for \texttt{pain} and \texttt{fever}. Results for the other symptoms across all subsets and training sizes are available in Appendix \ref{sec:app_PM_subsets}.

As seen in Table \ref{tab:comparison_models_brier_subsets_pain_fever}, \textbf{V-C-BN-text} reliably improves over \textbf{text-only} on the \textit{not present, mentioned} and \textit{not present, not mentioned} subsets, and displays varied performance on the \textit{present, not mentioned} subset. Performance is slightly worse on the \textit{present, mentioned} subset, but as seen in the Tables \ref{tab:comparison_models_average_precision} and \ref{tab:comparison_models_Brier}, \textbf{V-C-BN-text} still improves overall. (Note that the \textbf{text-only} model should be expected to make a good prediction for this subset, as the clinical note contains the required information.) These trends hold for the other symptoms and training sizes as well. Together, these results indicate that the BN provides \textbf{V-C-BN-text} with information about the prior distribution, which is biased towards the symptom(s) not being present.

The \textit{present, not mentioned} subset -- where the patient experiences a symptom which is not mentioned in the text -- is of particular interest, as the BN can provide valuable complementary information in this case. Table \ref{tab:brier_bn_nn_pnm} directly compares the performance of the BN and neural classifiers on this subset. As expected, the BN performs much better on this subset than the \textbf{text-only} classifiers. Intuitively, this occurs because the neural classifiers learn that a symptom is almost never present without being mentioned in the text, while the BN looks only at the other tabular data to form its prediction. The superior performance of the BN in this subset in turn should allow the combined \textbf{V-C-BN-text} model to improve more over the \textbf{text-only} baseline, as incorporating the BN's prediction helps correct for the missing information in the text. In Appendix \ref{sec:app_illustrative_example}, we break down a concrete example of \textbf{V-C-BN-text} providing a higher probability than \textbf{text-only} that \texttt{pain} is present when the symptom is not mentioned.

However, Table \ref{tab:comparison_models_brier_subsets_pain_fever} shows that while \textbf{V-C-BN-text} performs much better than the \textbf{text-only} classifiers on the \textit{present, not mentioned} subset for small training sizes (improving by a much larger margin than in other subsets), at larger training sizes this improvement breaks down for \texttt{pain} and \texttt{fever}. Table \ref{tab:brier_bn_nn_pnm} partly explains the particularly noticeable degradation in performance for \texttt{fever}: the BN is much worse at predicting the occurrence of \texttt{fever} in this subset than the other symptoms, leaving less room for it to improve the combined \textbf{V-C-BN-text} model. The general degradation is explained by the fact that, as the training size increases, the \textbf{text-only} classifiers become 
more confident in their predictions, making their contributions as virtual evidence weigh more heavily. This is detrimental in the \textit{present, not mentioned} subset where those \textbf{text-only} predictions are typically wrong.

As shown in Table \ref{tab:comparison_models_brier_subsets_pain_fever}, the consistency node \textbf{C-BN-text} significantly improves over the \textbf{text-only} classifiers for the \textit{present, not mentioned} subset for all training sizes, while the virtual evidence-only model \textbf{V-BN-text} often performs worse than the \textbf{text-only} classifiers. \textbf{V-C-BN-text}, using both virtual evidence and the consistency node, reliably improves over \textbf{V-BN-text}, indicating the consistency node's ability to help offset the fatal weakness of virtual evidence: confident (but wrong) virtual evidence can overwhelm the BN, leading to a more confident (and more wrong) final prediction. A more comprehensive analysis can be found in Appendix \ref{sec:app_PM_subsets}.

\begin{table*}[t]
  \caption{Brier scores ($\downdownarrows$) for our models on the present vs. mentioned subsets across various training sizes. The best model per training size and per symptom is highlighted in \textbf{bold}. Cases where a model outperforms \textbf{text-only} significantly are indicated by \textbf{*} ($p < 0.05$ in a one-sided Wilcoxon signed-rank test over 20 seeds).}
  \label{tab:comparison_models_brier_subsets_pain_fever}
\resizebox{\textwidth}{!}{
\begin{tabular}{lllcccccccc}
    \toprule
    & & & \multicolumn{8}{c}{Training size $n$} \\ \cmidrule{3-11}
    & & & \textbf{100} & \textbf{187} & \textbf{350} & \textbf{654} & \textbf{1223} & \textbf{2287} & \textbf{4278} & \textbf{8000}\\
    \midrule
    \textit{present,} & \texttt{pain} & \textbf{text-only} & \textbf{0.3809} & \textbf{0.2508} & \textbf{0.1941} & \textbf{0.1393} & \textbf{0.1252} & \textbf{0.103} & \textbf{0.0909} & \textbf{0.0706} \\
    \textit{mentioned} & & \textbf{C-BN-text} & 0.4679 & 0.3097 & 0.3086 & 0.2337 & 0.1912 & 0.1696 & 0.1489 & 0.1162 \\
    & & \textbf{V-BN-text} & 0.5569 & 0.401 & 0.3573 & 0.2387 & 0.1921 & 0.15 & 0.1247 & 0.0931 \\
    & & \textbf{V-C-BN-text} & 0.4157 & 0.28 & 0.2625 & 0.1915 & 0.1608 & 0.1354 & 0.1147 & 0.0885 \\
    \cmidrule{2-11}
    & \texttt{fever} & \textbf{text-only} & \textbf{0.7589} & \textbf{0.5334} & \textbf{0.4449} & \textbf{0.2503} & \textbf{0.1718} & \textbf{0.155} & \textbf{0.0998} & \textbf{0.0722} \\
    & & \textbf{C-BN-text} & 0.8994 & 0.6497 & 0.5861 & 0.3683 & 0.2582 & 0.2037 & 0.1579 & 0.1223 \\
    & & \textbf{V-BN-text} & 1.0834 & 0.7634 & 0.6551 & 0.3741 & 0.2486 & 0.2152 & 0.1294 & 0.0905 \\
    & & \textbf{V-C-BN-text} & 0.8151 & 0.5772 & 0.4998 & 0.2932 & 0.1965 & 0.1643 & 0.1076 & 0.0774 \\
    \midrule
    \textit{present,} & \texttt{pain} & \textbf{text-only} & 0.7786 & 0.8574 & 0.8387 & 0.8591 & 0.8822 & 0.8734 & 0.8711 & 0.8674 \\
    \textit{not mentioned} & & \textbf{C-BN-text} & *0.7251 & \textbf{*0.8035} & \textbf{*0.7825} & \textbf{*0.8102} & \textbf{*0.8236} & \textbf{*0.8236} & \textbf{*0.8206} & \textbf{*0.8178} \\
    & & \textbf{V-BN-text} & 0.8285 & 0.8962 & 0.8896 & 0.9142 & 0.9304 & 0.922 & 0.9188 & 0.9092 \\
    & & \textbf{V-C-BN-text} & \textbf{*0.724} & *0.8182 & *0.8038 & 0.8442 & *0.863 & 0.8695 & 0.8682 & 0.8679 \\
    \cmidrule{2-11}
    & \texttt{fever} & \textbf{text-only} & 1.6366 & 1.5965 & 1.5683 & 1.4084 & 1.39 & 1.3764 & 1.306 & 1.3218 \\
    & & \textbf{C-BN-text} & \textbf{*1.4653} & \textbf{*1.4837} & \textbf{*1.4503} & \textbf{1.3786} & \textbf{*1.3556} & \textbf{*1.3157} & \textbf{1.2763} & \textbf{*1.277} \\
    & & \textbf{V-BN-text} & 1.7423 & 1.7089 & 1.701 & 1.5679 & 1.5602 & 1.5568 & 1.5109 & 1.5409 \\
    & & \textbf{V-C-BN-text} & *1.511 & *1.5258 & *1.4959 & 1.4464 & 1.4325 & 1.3999 & 1.3855 & 1.4088 \\
    \midrule
    \textit{not present,} & \texttt{pain} & \textbf{text-only} & 0.0293 & 0.032 & 0.0305 & 0.0275 & 0.0169 & 0.0127 & 0.0103 & 0.0066 \\
    \textit{mentioned} & & \textbf{C-BN-text} & *0.0246 & *0.0253 & *0.0208 & *0.0184 & *0.0134 & *0.0107 & 0.0089 & 0.007 \\
    & & \textbf{V-BN-text} & \textbf{*0.0138} & \textbf{*0.0144} & \textbf{*0.0112} & \textbf{*0.009} & \textbf{*0.0055} & \textbf{*0.0043} & \textbf{*0.0035} & \textbf{*0.0025} \\
    & & \textbf{V-C-BN-text} & *0.0264 & *0.0258 & *0.0203 & *0.0145 & *0.009 & *0.0063 & *0.0051 & *0.0036 \\
    \cmidrule{2-11}
    & \texttt{fever} & \textbf{text-only} & 0.0929 & 0.1232 & 0.1025 & 0.1117 & 0.0691 & 0.0368 & 0.0318 & 0.019 \\
    & & \textbf{C-BN-text} & *0.0728 & *0.0821 & *0.0697 & *0.0678 & *0.0467 & *0.0307 & *0.0244 & 0.0183 \\
    & & \textbf{V-BN-text} & \textbf{*0.0278} & \textbf{*0.0361} & \textbf{*0.0263} & \textbf{*0.0329} & \textbf{*0.0195} & \textbf{*0.01} & \textbf{*0.0086} & \textbf{*0.0057} \\
    & & \textbf{V-C-BN-text} & *0.0724 & *0.078 & *0.0623 & *0.0507 & *0.0301 & *0.0193 & *0.0142 & *0.0107 \\
    \midrule
    \textit{not present,} & \texttt{pain} & \textbf{text-only} & 0.021 & 0.0164 & 0.0147 & 0.0131 & 0.0097 & 0.0098 & 0.0094 & 0.0081 \\
    \textit{not mentioned} & & \textbf{C-BN-text} & 0.0196 & *0.015 & *0.013 & 0.0106 & 0.0091 & 0.0089 & 0.0086 & 0.0082 \\
    & & \textbf{V-BN-text} & \textbf{*0.011} & \textbf{*0.0083} & \textbf{*0.0059} & \textbf{*0.0047} & \textbf{*0.0037} & \textbf{*0.0041} & \textbf{*0.0039} & \textbf{*0.0039} \\
    & & \textbf{V-C-BN-text} & 0.0205 & *0.0147 & *0.0117 & *0.0081 & *0.0063 & *0.0058 & *0.0053 & *0.0049 \\
    \cmidrule{2-11}
    & \texttt{fever} & \textbf{text-only} & 0.0289 & 0.0293 & 0.0265 & 0.0527 & 0.05 & 0.0388 & 0.0427 & 0.029 \\
    & & \textbf{C-BN-text} & 0.0367 & 0.0271 & 0.0278 & *0.0356 & *0.0359 & *0.0328 & *0.0331 & 0.0276 \\
    & & \textbf{V-BN-text} & \textbf{*0.0096} & \textbf{*0.0088} & \textbf{*0.0072} & \textbf{*0.0163} & \textbf{*0.017} & \textbf{*0.0122} & \textbf{*0.0121} & \textbf{*0.0076} \\
    & & \textbf{V-C-BN-text} & 0.0311 & *0.0223 & *0.0213 & *0.0249 & *0.0246 & *0.0211 & *0.0186 & *0.0138 \\
    \bottomrule
    \bottomrule
\end{tabular}}
\end{table*}

\begin{table*}[t]
  \caption{Brier scores ($\downdownarrows$) for BN-only vs. text-only on the \textit{present, not mentioned} subset across various training sizes. Note that BN-only performs much better than text-only for all symptoms save fever.}
  \label{tab:brier_bn_nn_pnm}
\resizebox{\textwidth}{!}{
\begin{tabular}{llcccccccc}
    \toprule
    & & \multicolumn{8}{c}{Training size $n$} \\ \cmidrule{3-10}
    & & \textbf{100} & \textbf{187} & \textbf{350} & \textbf{654} & \textbf{1223} & \textbf{2287} & \textbf{4278} & \textbf{8000}\\
    \midrule
    \texttt{dysp} & \textbf{BN-only} & 0.2892 & 0.2814 & 0.2608 & 0.2632 & 0.2582 & 0.2555 & 0.2579 & 0.2612 \\
    & \textbf{text-only} & 0.8879 & 0.8893 & 0.8958 & 0.9192 & 0.8972 & 0.8517 & 0.8501 & 0.847 \\
    \midrule
    \texttt{cough} & \textbf{BN-only} & 0.2019 & 0.1886 & 0.177 & 0.1793 & 0.1791 & 0.179 & 0.1803 & 0.1761 \\
    & \textbf{text-only} & 0.5749 & 0.642 & 0.6555 & 0.644 & 0.66 & 0.6452 & 0.6715 & 0.6648 \\
    \midrule
    \texttt{pain} & \textbf{BN-only} & 0.5822 & 0.5994 & 0.5925 & 0.5954 & 0.5975 & 0.5995 & 0.5951 & 0.5954 \\
    & \textbf{text-only} & 0.7786 & 0.8574 & 0.8387 & 0.8591 & 0.8822 & 0.8734 & 0.8711 & 0.8674 \\
    \midrule
    \texttt{nasal} & \textbf{BN-only} & 0.2982 & 0.2931 & 0.2868 & 0.2868 & 0.283 & 0.2813 & 0.2751 & 0.2827 \\
    & \textbf{text-only} & 0.8839 & 0.9129 & 0.9198 & 0.9324 & 0.9298 & 0.899 & 0.9087 & 0.8948 \\
    \midrule
    \texttt{fever} & \textbf{BN-only} & 1.1247 & 1.0954 & 1.098 & 1.0774 & 1.0818 & 1.0847 & 1.0908 & 1.0905 \\
    & \textbf{text-only} & 1.6366 & 1.5965 & 1.5683 & 1.4084 & 1.39 & 1.3764 & 1.306 & 1.3218 \\
    \bottomrule
    \bottomrule
\end{tabular}}
\end{table*}

\subsection{Handling shifts in text data distribution} \label{sec:data_shift}

The performance gaps between the combined models and the \textbf{text-only} classifier in Table \ref{tab:comparison_models_average_precision} and Table \ref{tab:comparison_models_Brier} are often 
small, which can be attributed to the limited information gap between the tabular data and what is written in the text notes in the SimSUM dataset. Furthermore, our analysis of performance on the \textit{present, not mentioned} subset shows the potential of our method to correct for faulty predictions of the \textbf{text-only} classifiers when information is missing from the text.

To investigate how our method performs when more information is missing from the text, leaving more room for improvement over the \textbf{text-only} model, we create a new version of the test dataset, which we call $\mathcal{X}_{test}^*$, containing manipulated notes. In this test set, the tabular variables remain the same, but we randomly mask out sentences from the notes describing a symptom. For this, we use the annotated symptom spans that were released together with the SimSUM dataset \citep{SimSUM}. For every note in the test set, we go over each phrase that describes any of the symptoms, and drop the sentence containing that phrase with a 50\% probability. For example, the full sentence ``Patient presents with low-grade fever and significant nasal symptoms'' might be dropped to remove the mention of ``low-grade fever'' or of ``significant nasal symptoms''. Note that this technique increases the gap between the information contained in the tabular portion of the data (patient background variables causing certain symptoms), and the information contained in the text portion of the data (descriptions of these symptoms in the clinical notes).

We then used the original \textbf{BN-only} and \textbf{text-only} classifiers (which were trained using the original train set $\mathcal{X}_{train}$ with non-manipulated notes) to evaluate performance on this new test set $\mathcal{X}_{test}^*$. We also use the original consistency node, whose weights were set based on the original train set $\mathcal{X}_{train}$. Tables \ref{tab:data_shift_average_precision_pf} and \ref{tab:data_shift_Brier_pf} show the average precision and Brier scores, respectively, of these models on $\mathcal{X}_{test}^*$, the set of out-of-distribution samples, for the symptoms \texttt{pain} and \texttt{fever}. These results demonstrate larger improvements than seen before in Tables \ref{tab:comparison_models_average_precision} and \ref{tab:comparison_models_Brier}, due to the added room for improvement in the \textbf{text-only} classifiers as a result of the text data shift in $\mathcal{X}_{test}^*$. This pattern also holds for the other symptoms. The full results can be found in Appendix \ref{sec:app_data_shift_results}.
The \textbf{V-C-BN-text} model significantly improves over the \textbf{text-only} baseline for almost all training sizes and all symptoms. This shows that by including the BN and its background knowledge, the \textbf{V-C-BN-text} model can fill in gaps of missing information in the text, allowing it to effectively handle shifts in text data distribution (compared to the training phase) during inference.

\begin{table*}[t]
  \caption{Average precision ($\upuparrows$) for the predictions of our models over the test set $\mathcal{X}_{test}^*$ containing \textbf{manipulated text notes}. The best model per training size and per symptom is highlighted in \textbf{bold}. Cases where a model outperforms \textbf{text-only} significantly are indicated by \textbf{*} ($p < 0.05$ in a one-sided Wilcoxon signed-rank test over 20 seeds).}
  \label{tab:data_shift_average_precision_pf}
\resizebox{\textwidth}{!}{
\begin{tabular}{llcccccccc}
    \toprule
    & & \multicolumn{8}{c}{Training size $n$} \\ \cmidrule{3-10}
    & & \textbf{100} & \textbf{187} & \textbf{350} & \textbf{654} & \textbf{1223} & \textbf{2287} & \textbf{4278} & \textbf{8000}\\
    \midrule
    \texttt{pain} & \textbf{text-only} & 0.5357 & 0.6195 & 0.6575 & 0.6946 & 0.722 & 0.7349 & 0.7422 & 0.7516 \\
    & \textbf{C-BN-text} & \textbf{*0.5467} & \textbf{*0.6262} & \textbf{*0.669} & *0.7031 & *0.7285 & *0.7404 & *0.7493 & *0.7595 \\
    & \textbf{V-BN-text} & 0.5176 & 0.6106 & 0.6567 & \textbf{*0.714} & \textbf{*0.7473} & \textbf{*0.764} & \textbf{*0.775} & \textbf{*0.7868} \\
    & \textbf{V-C-BN-text} & 0.538 & 0.6228 & *0.6632 & *0.7065 & *0.737 & *0.7528 & *0.7613 & *0.7723 \\
    \cline{2-10}
    & \textbf{change vs. baseline} & +1.1\% & +0.67\% & +1.15\% & +1.94\% & +2.53\% & +2.91\% & +3.27\% & +3.52\% \\
    \midrule
    \texttt{fever} & \textbf{text-only} & 0.6023 & 0.6524 & 0.6875 & 0.7361 & 0.7714 & 0.7909 & 0.8072 & 0.8161 \\
    & \textbf{C-BN-text} & 0.5975 & *0.6562 & *0.6943 & *0.7424 & *0.7761 & *0.7981 & *0.8132 & *0.8243 \\
    & \textbf{V-BN-text} & \textbf{0.6047} & \textbf{*0.6782} & \textbf{*0.7202} & \textbf{*0.7746} & \textbf{*0.8086} & \textbf{*0.8269} & \textbf{*0.8439} & \textbf{*0.8543} \\
    & \textbf{V-C-BN-text} & 0.6034 & *0.6671 & *0.7044 & *0.7572 & *0.7915 & *0.8104 & *0.829 & *0.839 \\
    \cline{2-10}
    & \textbf{change vs. baseline} & +0.23\% & +2.58\% & +3.27\% & +3.86\% & +3.72\% & +3.6\% & +3.67\% & +3.83\% \\
    \bottomrule
    \bottomrule
\end{tabular}}
\end{table*}

\begin{table*}[t]
  \caption{Brier scores ($\downdownarrows$) for the predictions of our models over the test set $\mathcal{X}_{test}^*$ containing \textbf{manipulated text notes}. The best model per training size and per symptom is highlighted in \textbf{bold}. Cases where a model outperforms \textbf{text-only} significantly are indicated by \textbf{*} ($p < 0.05$ in a one-sided Wilcoxon signed-rank test over 20 seeds).}
  \label{tab:data_shift_Brier_pf}
\resizebox{\textwidth}{!}{
\begin{tabular}{llcccccccc}
    \toprule
    & & \multicolumn{8}{c}{Training size $n$} \\ \cmidrule{3-10}
    & & \textbf{100} & \textbf{187} & \textbf{350} & \textbf{654} & \textbf{1223} & \textbf{2287} & \textbf{4278} & \textbf{8000}\\
    \midrule
    \texttt{pain} & \textbf{text-only} & 0.0955 & 0.0879 & 0.0793 & 0.0753 & 0.0698 & 0.0665 & 0.0657 & 0.0621 \\
    & \textbf{C-BN-text} & 0.0982 & 0.0867 & 0.082 & 0.0754 & 0.0703 & 0.0679 & 0.0662 & 0.0628 \\
    & \textbf{V-BN-text} & 0.1043 & 0.0933 & 0.0869 & 0.0773 & 0.0733 & 0.0694 & 0.067 & 0.0639 \\
    & \textbf{V-C-BN-text} & \textbf{0.0949} & \textbf{*0.0853} & \textbf{0.0789} & \textbf{*0.072} & \textbf{0.0683} & \textbf{0.0661} & \textbf{0.0641} & \textbf{0.0618} \\
    \cline{2-10}
    & \textbf{change vs. baseline} & -0.05\% & -0.26\% & -0.04\% & -0.33\% & -0.15\% & -0.04\% & -0.16\% & -0.03\% \\
    \midrule
    \texttt{fever} & \textbf{text-only} & 0.3071 & 0.2887 & 0.2695 & 0.2613 & 0.2388 & 0.2231 & 0.2134 & 0.2041 \\
    & \textbf{C-BN-text} & 0.3029 & *0.2743 & *0.2612 & *0.2401 & *0.2222 & *0.2102 & *0.2012 & *0.196 \\
    & \textbf{V-BN-text} & 0.3233 & 0.2878 & 0.2736 & *0.2437 & *0.2251 & 0.2171 & 0.2051 & 0.2036 \\
    & \textbf{V-C-BN-text} & \textbf{*0.294} & \textbf{*0.2671} & \textbf{*0.2515} & \textbf{*0.2306} & \textbf{*0.2131} & \textbf{*0.2034} & \textbf{*0.1949} & \textbf{*0.1929} \\
    \cline{2-10}
    & \textbf{change vs. baseline} & -1.31\% & -2.16\% & -1.8\% & -3.07\% & -2.57\% & -1.97\% & -1.85\% & -1.12\% \\
    \bottomrule
    \bottomrule
\end{tabular}}
\end{table*}

\subsection{Key takeaways} \label{sec:takeaways}


Our results show that the methods combining the BN with neural classifiers outperform the \textbf{text-only} and \textbf{concat-tab-text} baselines, with the \textbf{V-C-BN-text} model performing the best overall. From our analysis of present vs. mentioned subsets, we found that \textbf{V-C-BN-text} reliably improves over the \textbf{text-only} classifiers in cases where the symptom is not present, indicating that incorporating the predictions of the BN better accounts for the true prior (which is biased towards a symptom not being present).

Furthermore, we found that for symptoms and training sizes where the BN performs well relative to the \textbf{text-only} classifiers, the largest improvements of the combined \textbf{V-C-BN-text} model come from cases where the symptom is present but not mentioned in the text. This hints at the ability of the BN to fill in the gaps where information is missing from the text. This ability is further demonstrated in a data shift experiment, where information is missing from the text at a higher rate at inference time than during training. In this setting, the combined models outperform the \textbf{text-only} classifiers by a larger margin, indicating that the BN is able to help correct for the missing information.

Finally, the Brier scores of \textbf{V-C-BN-text} relative to \textbf{V-BN-text} indicate the consistency node's ability to improve calibration over virtual evidence alone, while retaining strong overall accuracy. The difference in Brier scores are especially notable when the symptom is missing from the text (see \textit{present, not mentioned} in Table \ref{tab:comparison_models_brier_subsets_pain_fever}), where virtual evidence is particularly susceptible to confident, incorrect predictions from the neural classifiers.



\section{Conclusion} \label{sec:conclusion}

In this work, we introduced the concept of patient-level information extraction that leverages both the structured tabular features in a patient’s EHR and the unstructured clinical notes describing the patient’s symptoms. By augmenting \textit{virtual evidence} with a \textit{consistency node}, we achieved interpretable integration of a Bayesian network with neural text classifiers, enabling coherent and probabilistic fusion of tabular and textual information. Our method resulted in better-calibrated final predictions for the target variables that take into account the true prior encoded in the Bayesian network. At the same time, it highlighted the potential of the Bayesian network to correct for abnormal cases of missing textual data. Our method proved most effective for middling training sizes where the BN approaches its optimal performance prior to the neural classifiers, making it particularly appealing for use-cases where training data is limited.

While the current work focused on the specific use-case of predicting patient-level symptoms from tabular data and text, we foresee a broader use of our method in the future. First, any node in the Bayesian network may be the target of information extraction, as long as the text contains some information about this feature. Second, and more broadly, either of the two modalities (tabular and text) may be swapped out for any other. Virtual evidence and the consistency node are flexible, only expecting probabilities at the input, without assuming any particular underlying method with which these probabilities are obtained. For example, another application of our combined consistency method could be the automated extraction of information from X-Ray images, where predictions of an image classifier that detects radiology features can be easily combined with a Bayesian network that includes tabular background information on the patient (such as age, previous diagnoses, etc.). Furthermore, while virtual evidence requires a Bayesian network, there is no requirement that a neural network be used to model the other modality. The consistency node is completely agnostic to the choice of models and can even be used without a Bayesian network.\footnote{However, we do find that having a Bayesian network as the tabular data model would be an asset in many medical use-cases, thanks to the implicit inclusion of expert knowledge.} Finally, with minor adjustments to the consistency node, it becomes possible to include more than two modalities, while virtual evidence is inherently able to handle evidence from multiple modalities for the same node of a Bayesian network. In the previous example, radiology reports might be included as a third modality along with X-ray images and tabular background information.

Ultimately, the consistency node is a promising approach that provides an interpretable fusion of arbitrary models and modalities, while its compatibility with virtual evidence makes it particularly suitable for integration with Bayesian networks.

\section{Limitations} \label{sec:limitations}

Our method has several limitations. First, we interpret the probabilities at the output of the neural classifier as if they are a reflection of its confidence on the presence of the symptom in the text. However, this is not necessarily the case, as neural classifiers are known to have issues with calibration \citep{calibration}. Still, this is offset by the consistency node, which improves calibration compared to using virtual evidence alone
, as evidenced by the elevated Brier scores. 


Second, we make strong assumptions on the types of conditional distributions that are learned in the BN. In our case, these assumptions match up perfectly with the true data generating process of the data as described in \citet{SimSUM}. However, in a realistic setting, one would not have access to the true type of probability distribution for each variable in the network, and would instead need to consult an expert. 

Third, it can be very challenging to come up with a DAG structure that accurately captures reality. To mitigate this, one could work with a panel of experts who iteratively improve the DAG. Furthermore, future work can focus on using (partial) structure learning algorithms \citep{BN_structure_learning} to learn the DAG from the data, filling in the gaps where experts are unsure, while still asking experts to validate the final DAG. Note that while the inclusion of the BN in our method might limit its generalization to broader contexts, we explicitly choose to trade in this flexibility for interpretability and expert input.

Finally, and related to the previous point, we only validated our method on a single simulated use-case. While this shows the merit of our method as a proof-of-concept, future work should focus on putting the theory into practice and applying our method to a more realistic and challenging dataset. To this end, the MIMIC-III \citep{mimic3} and MIMIC-IV \citep{mimic4} datasets come to mind. 


\bmhead{Data and Code Availability}
This paper uses the SimSUM dataset, which is freely available on Github \citep{SimSUM}. Our code is available at \url{https://github.com/AdrickTench/patient-level-IE}. 

\bmhead{Institutional Review Board (IRB)}
Since we conduct all experiments on a simulated dataset, our research did not require IRB approval.

\bmhead{Acknowledgements}
Paloma Rabaey's research is funded by the Research Foundation Flanders (FWO Vlaanderen) with grant number 1170124N. This research has also received funding from the Flemish government under the “Onderzoeksprogramma Artificiele Intelligentie (AI) Vlaanderen” programme. The authors thank Robin Manhaeve and Jaron Maene for their valuable insights during the conception of the consistency node method and their helpful feedback on early versions of the manuscript. 

\bibliography{references}


\begin{appendices}

\newpage
\section{Training details}\label{sec:app_training}

\subsection{Bayesian network} \label{sec:app_training_BN}

To model the tabular portion of the data, we use a Bayesian network. A Bayesian network is defined by a Directed Acyclic Graph (DAG), which models the relations between the variables. This DAG also prescribes how the joint distribution factorizes into conditional probability distributions, one for each variable conditional on its parents. In SimSUM, both the DAG and the probability distributions are defined by an expert, together forming a Bayesian network from which the tabular data is sampled.
In a realistic setting, we cannot assume that we know the full data generating process. Instead, we assume that we can consult an expert to tell us how the variables are related, giving rise to the DAG in Figure \ref{fig:consistency_overview}, but that we need to learn the exact conditional probability distributions from data. 

Formally, we model the tabular portion of the data by learning the probability distribution in Equation \ref{eq:tab_distr}, where we abbreviated some of the variable names for ease of presentation. 

\begin{footnotesize}
\begin{align}
    &\mathcal{P}_{tab}(\text{\texttt{Asthma, Smoking, COPD, Hay, Season, Pneu, Cold}} \nonumber \\
    &\text{\texttt{Dysp, Cough, Pain, Nasal, Fever, Antibio, \#Days}}) \nonumber\\
    &= 
    \mathcal{P}(\texttt{Asthma}) 
    \mathcal{P}(\texttt{Smoking}) 
    \mathcal{P}(\texttt{COPD} \mid \texttt{Smoking})
    \mathcal{P}(\texttt{Season})\nonumber\\
    &\mathcal{P}(\texttt{Hay})
    \mathcal{P}(\texttt{Cold} \mid \texttt{Season}) 
    \mathcal{P}(\texttt{Pneu} \mid \texttt{Asthma, COPD, Season}) 
    \nonumber\\
    &\mathcal{P}(\texttt{Dysp} \mid \texttt{Asthma, Smoking, COPD, Pneu, Hay}) \nonumber\\
    &\mathcal{P}(\texttt{Cough} \mid \texttt{Asthma, Smoking, COPD, Pneu, Cold}) \nonumber\\
    &\mathcal{P}(\texttt{Pain} \mid \texttt{Cough, Pneu, COPD, Cold}) 
    \mathcal{P}(\texttt{Fever} \mid \texttt{Pneu, Cold}) \nonumber\\
    &\mathcal{P}(\texttt{Nasal} \mid \texttt{Cold, Hay}) 
    \mathcal{P}(\texttt{Antibio} \mid \texttt{Dysp, Cough, Pain, Fever}) \nonumber \\
    &\mathcal{P}(\texttt{\#Days} \mid \texttt{Antibio, Dysp, Cough, Pain, Fever, Nasal}) \nonumber\\
\label{eq:tab_distr}
\end{align}
\end{footnotesize}

In SimSUM, the conditional probability distributions are parameterized in various ways: (i) conditional probability tables (CPTs) for the variables \texttt{Asthma}, \texttt{Smoking}, \texttt{COPD}, \texttt{Hay fever}, \texttt{Season}, \texttt{Pneumonia}, \texttt{Common cold} and \texttt{Fever}, (ii) Noisy-OR distributions for the symptoms \texttt{Dyspnea}, \texttt{Cough}, \texttt{Pain}, and \texttt{Nasal}, (iii) a logistic regression model and two Poisson regression models for the variables \texttt{Antibiotics} and \texttt{\#Days}, respectively. 

To obtain the full probability distribution {\small$\mathcal{P}_{tab}(\text{\texttt{Asthma, Smoking, ..., Antibiotics,}} \allowbreak \text{\texttt{\#Days}})$}, we can learn the parameters for each of these conditional distributions independently, by training them on the tabular portion of the data $\mathcal{X}_{train}$. To this end, we follow the approach outlined in SimSUM \citep{SimSUM}. In short, all parameters are learned through Maximum Likelihood Estimation, where the exact likelihood that is optimized depends on the particular parametrization approach. We manually tune the hyperparameters (learning rate and number of epochs) for each training set size, increasing the number of epochs and learning rate for smaller training sets. We keep the batch size fixed at 50. 

When all parameters have been learned, we turn these distributions into CPTs by evaluating them for each combination of child and parent values, as described in SimSUM \citep{SimSUM}. This allows us to do exact inference over the tabular evidence in the learned Bayesian network through variable elimination \citep{koller2009probabilistic}. 


\subsection{Neural text classifier} \label{app:neural_text}

We follow the approach of \citet{SimSUM} for training the neural text classifier. There is one hidden layer of size $256$, followed by a ReLU activation. To deal with the varying training set sizes, we tune the optimal number of epochs with early stopping in a 5-fold cross-validation loop for each training set, with patience of 10 and tolerance of $10^{-3}$ on the cross-entropy loss over the validation set (with the maximum set to 200 epochs). We then take the median of the number of epochs at which early stopping was applied for each of these cross-validation splits, and retrain with the full training set afterwards for that number of epochs. The other hyperparameters are fixed as follows: a batch size of 50, a learning rate of 0.0005, weight decay (L2 regularization) of $10^{-5}$ and no dropout. 

\subsection{Multimodal baseline: Concat-tab-text}\label{app:multi_modal_base}

We use the implementation of the neural-text-tab baseline as described in \citet{SimSUM}. According to this implementation, the tabular features are transformed into a vector representation. We use a one-hot encoding for the categorical (binary) features, and normalize the \texttt{\#Days} feature using a StandardScaler. This tabular feature representation (vector of shape $9$) is concatenated with the text embedding representation (vector of shape $768$) and then fed into the exact same architecture as was used for the text-only baseline (only its input layer is updated from size $768$ to size $777$). Accordingly, we use the exact same cross-validation strategy and hyperparameters to train the model as described in Section \ref{app:neural_text}.

\section{Consistency node numerical example}\label{sec:numerical_example}

To illustrate the functionality of the consistency node, we provide a simple numerical example. We provide ground truth labels and probabilities from a text classifier and BN for four fictional patients in Table \ref{tab:numerical_example_symptom_probs}.

\begin{table}[!htbp]
    \centering
    \begin{tabular}{lccc}
        \hline
         & GT label & $\mathcal{P}(T \mid note)$ & $\mathcal{P}(BN \mid tab)$ \\
        \hline
        1 & \texttt{yes} & 0.1 & 0.7 \\
        2 & \texttt{no}  & 0.1 & 0.2 \\
        3 & \texttt{no}  & 0.1 & 0.6 \\
        4 & \texttt{yes} & 0.9 & 0.6 \\
        \hline
    \end{tabular}
    \caption{Initial symptom probabilities predicted for four samples by the text classifier ($\mathcal{P}(T \mid note)$) and the BN ($\mathcal{P}(BN \mid tab)$). GT label indicates the ground truth label for the symptom in the training set.}
    \label{tab:numerical_example_symptom_probs}
\end{table}

Table \ref{tab:numerical_example_agreement} shows the agreement between the text classifier and BN across these 4 patients. This agreement is calculated through Equation \ref{eq:consistency_weight_fractions} in Section \ref{sec:consistency_node}. For example, the first row of Table \ref{tab:numerical_example_agreement} is calculated using the probabilities provided by the text classifier and BN that the symptom is \texttt{no} in both cases of the ground truth label. Concretely, the probabilities from the cases where the ground truth is \texttt{no} are $(1 - 0.1) \times (1 - 0.2) + (1 - 0.9) \times (1 - 0.6)$ (yielding a weight of $W\{S = \texttt{no}, BN = \texttt{no}, T = \texttt{no}\}$ = $1.08$) while the probabilities where the ground truth is \texttt{yes} are $(1 - 0.1) \times (1 - 0.7) + (1 - 0.9) \times (1 - 0.6)$ (yielding a weight of $W\{S = \texttt{yes}, BN = \texttt{no}, T = \texttt{no}\}$ = $0.31$). These weights are then normalized by dividing by the sum for the row ($W\{S = \texttt{no}, BN = \texttt{no}, T = \texttt{no}\}+W\{S = \texttt{yes}, BN = \texttt{no}, T = \texttt{no}\}$ = $1.39$), yielding the probabilities in the first row of Table \ref{tab:numerical_example_agreement}.

\begin{table}[!htbp]
    \centering
    \begin{tabular}{cc|cc}
        \hline
        T & BN & $\mathcal{P}(C = \texttt{no} \mid T, BN)$ & $\mathcal{P}(C = \texttt{yes} \mid T, BN)$ \\
        \hline
        \texttt{no} & \texttt{no} & 0.78 & 0.22 \\
        \texttt{no} & \texttt{yes} & 0.51 & 0.49 \\
        \texttt{yes} & \texttt{no} & 0.24 & 0.76 \\
        \texttt{yes} & \texttt{yes} & 0.12 & 0.88 \\
        \hline
    \end{tabular}
    \caption{Consistency node probabilities $\mathcal{P}(C \mid T, BN)$ obtained from predictions in Table \ref{tab:numerical_example_symptom_probs}.}
    \label{tab:numerical_example_agreement}
\end{table}

Given the consistency node probabilities in Table \ref{tab:numerical_example_agreement}, the final probability for the presence of a symptom, $\mathcal{P}(C = \texttt{yes} \mid tab, note)$, can be calculated using the probabilities in \texttt{yes} column, following Equation \ref{eq:consistency_sum} in Section \ref{sec:consistency_node}. For example, given probabilities of $0.1$ from the text classifier and $0.8$ from the BN, the final probability is calculated as follows:

\begin{align}
\mathcal{P}(C = \texttt{yes} \mid tab, note) 
&= ((1-0.1) \times (1-0.8) \times 0.22) + ((1-0.1) \times 0.8 \times 0.49) \notag \\
&\quad + (0.1 \times (1-0.8) \times 0.76) + (0.1 \times 0.8 \times 0.88) \approx 0.48 \notag 
\end{align}

\section{Mentions label construction}\label{sec:mentions_label}

The SimSUM dataset does not include labels indicating if a symptom is mentioned in the clinical note. We created these labels from the information available in SimSUM, augmented by manual annotation. The labels we used are made available on our Github repository, along with the rest of the code, at \url{https://github.com/AdrickTench/patient-level-IE}.

SimSUM includes two key pieces of information that we used to create the mentions labels: (1) a label indicating if the LLM that generated the clinical note was instructed to mention (or not mention) the symptom, and (2) pre-identified span annotations indicating portions of the clinical note that mention the symptom. In cases where (1) and (2) agree, we automatically generated the corresponding label. Disagreements between (1) and (2) were resolved through manual annotation by one of the authors.

Such disagreements can be attributed to either a failure of the LLM to follow the instructions provided in its prompt, or a failure of the span annotation process. Because of the possibility that both (1) and (2) failed on the same symptom and note but were automatically accepted as the true label, and the possibility of human error in the manual annotation process, we cannot guarantee that our mentions labels are completely accurate. However, we expect that inaccuracies are exceptional and the labels are correct in the overwhelming majority of cases.

\section{Extended results}

\subsection{Results for all symptoms}\label{sec:app_all_symptom_results}

We report the average precision (Table \ref{tab:comparison_models_average_precision_all}) and Brier scores (Table \ref{tab:comparison_models_Brier_all}) of our models for all symptoms, as calculated on $\mathcal{X}_{test}$ over 20 seeds for various training sizes $n$. We also report the overall mean of each metric over all five symptoms. We find that \textbf{V-C-BN-text} performs the best overall, providing the best Brier scores and a statistically significant improvement in average precision. While the virtual-evidence-only model \textbf{V-BN-text} sometimes provides slightly better improvements to average precision, it does not yield statistically significant improvements to the Brier score as reliably as \textbf{V-C-BN-text}.

\begin{table*}[t]
  \caption{Average precision ($\upuparrows$) for the predictions of our models over $\mathcal{X}_{test}$, averaged over 20 seeds for various training sizes $n$. The best model per training size and per symptom is highlighted in \textbf{bold}. The best baseline model for each class is \underline{underlined}. Cases where a model outperforms the best baseline model significantly are indicated by \textbf{*} ($p < 0.05$ in a one-sided Wilcoxon signed-rank test over 20 seeds).}
  \label{tab:comparison_models_average_precision_all}
\resizebox{\textwidth}{!}{
\begin{tabular}{llcccccccc}
    \toprule
    & & \multicolumn{8}{c}{Training size $n$} \\ \cmidrule{3-10}
    & & \textbf{100} & \textbf{187} & \textbf{350} & \textbf{654} & \textbf{1223} & \textbf{2287} & \textbf{4278} & \textbf{8000}\\
    \midrule
    \texttt{dysp} & \textbf{BN-only} & 0.7625 & 0.7644 & 0.7794 & 0.7937 & 0.7972 & 0.7981 & 0.7981 & 0.7989 \\
    & \textbf{text-only} & \underline{0.9246} & \underline{0.9482} & \underline{0.9578} & \underline{0.967} & 0.9731 & 0.9798 & 0.9831 & \underline{0.9878} \\
    & \textbf{Concat-text-tab} & 0.9127 & 0.9398 & 0.9533 & 0.9657 & \underline{0.9737} & \underline{0.9801} & \underline{0.9841} & 0.987 \\
    \cline{2-10}
    & \textbf{C-BN-text} & *0.9258 & 0.948 & 0.957 & 0.9669 & 0.9736 & 0.9785 & 0.9817 & 0.9875 \\
    & \textbf{V-BN-text} & 0.9186 & 0.9392 & 0.9472 & 0.9665 & 0.9741 & 0.9806 & 0.9841 & *0.9882 \\
    & \textbf{V-C-BN-text} & \textbf{*0.928} & \textbf{*0.9519} & \textbf{*0.96} & \textbf{*0.9701} & \textbf{*0.9759} & \textbf{*0.9819} & \textbf{*0.9853} & \textbf{*0.9892} \\
    \cline{2-10}
    & \textbf{change vs. baseline} & +0.35\% & +0.36\% & +0.22\% & +0.31\% & +0.22\% & +0.18\% & +0.12\% & +0.13\% \\
    \midrule
    \texttt{cough} & \textbf{BN-only} & 0.7568 & 0.7718 & 0.7844 & 0.7898 & 0.7929 & 0.7946 & 0.7947 & 0.7942 \\
    & \textbf{text-only} & \underline{0.902} & \underline{0.9452} & 0.9571 & 0.9689 & 0.978 & 0.9826 & 0.9863 & 0.989 \\
    & \textbf{Concat-text-tab} & 0.8866 & 0.9393 & \underline{0.9582} & \underline{0.9705} & \underline{0.9794} & \underline{0.9851} & \underline{0.9886} & \underline{0.9908} \\
    \cline{2-10}
    & \textbf{C-BN-text} & *0.9086 & *0.9483 & *0.9607 & 0.9701 & 0.9772 & 0.9813 & 0.9858 & 0.988 \\
    & \textbf{V-BN-text} & 0.8988 & 0.9431 & 0.9578 & *0.9727 & \textbf{*0.9825} & \textbf{*0.9864} & \textbf{*0.9896} & \textbf{*0.9918} \\
    & \textbf{V-C-BN-text} & \textbf{*0.916} & \textbf{*0.9549} & \textbf{*0.9647} & \textbf{*0.9743} & *0.9821 & 0.9857 & 0.989 & *0.9914 \\
    \cline{2-10}
    & \textbf{change vs. baseline} & +1.4\% & +0.97\% & +0.65\% & +0.38\% & +0.3\% & +0.13\% & +0.1\% & +0.1\% \\
    \midrule
    \texttt{pain} & \textbf{BN-only} & 0.3197 & 0.3277 & 0.3409 & 0.3464 & 0.35 & 0.3509 & 0.3517 & 0.3515 \\
    & \textbf{text-only} & \textbf{\underline{0.6181}} & \underline{0.7252} & \underline{0.7648} & \underline{0.804} & \underline{0.825} & 0.8377 & 0.8471 & 0.8608 \\
    & \textbf{Concat-text-tab} & 0.5091 & 0.6537 & 0.7422 & 0.7858 & 0.8183 & \underline{0.8412} & \underline{0.8559} & \underline{0.868} \\
    \cline{2-10}
    & \textbf{C-BN-text} & 0.6135 & \textbf{*0.7271} & \textbf{*0.7723} & *0.8124 & *0.8323 & 0.8435 & 0.8521 & 0.8653 \\
    & \textbf{V-BN-text} & 0.5667 & 0.7027 & 0.7532 & \textbf{*0.8194} & \textbf{*0.8463} & \textbf{*0.8598} & \textbf{*0.8699} & \textbf{*0.8826} \\
    & \textbf{V-C-BN-text} & 0.6022 & 0.7244 & 0.7673 & *0.8146 & *0.8375 & *0.8511 & *0.8606 & *0.8738 \\
    \cline{2-10}
    & \textbf{change vs. baseline} & -0.46\% & +0.19\% & +0.76\% & +1.54\% & +2.13\% & +1.86\% & +1.4\% & +1.46\% \\
    \midrule
    \texttt{nasal} & \textbf{BN-only} & 0.6337 & 0.6357 & 0.6341 & 0.6303 & 0.634 & 0.6324 & 0.6328 & 0.634 \\
    & \textbf{text-only} & \underline{0.9512} & \underline{0.9584} & 0.9657 & 0.9706 & 0.9743 & 0.9779 & 0.98 & 0.9816 \\
    & \textbf{Concat-text-tab} & 0.9444 & 0.9579 & \underline{0.9666} & \underline{0.973} & \underline{0.9777} & \underline{0.9818} & \underline{0.9847} & \underline{0.9869} \\
    \cline{2-10}
    & \textbf{C-BN-text} & \textbf{*0.9567} & *0.9624 & 0.9686 & \textbf{*0.9769} & *0.9799 & 0.9817 & 0.9841 & 0.9847 \\
    & \textbf{V-BN-text} & 0.9508 & *0.9609 & 0.9664 & *0.976 & \textbf{*0.9819} & \textbf{*0.9861} & \textbf{*0.9879} & \textbf{*0.9885} \\
    & \textbf{V-C-BN-text} & *0.9547 & \textbf{*0.9628} & \textbf{*0.9699} & *0.9757 & *0.9798 & *0.9836 & *0.9858 & 0.9869 \\
    \cline{2-10}
    & \textbf{change vs. baseline} & +0.54\% & +0.45\% & +0.33\% & +0.39\% & +0.43\% & +0.43\% & +0.31\% & +0.16\% \\
    \midrule
    \texttt{fever} & \textbf{BN-only} & 0.4792 & 0.4983 & 0.5167 & 0.5309 & 0.5398 & 0.5437 & 0.5465 & 0.5474 \\
    & \textbf{text-only} & \textbf{\underline{0.6905}} & \underline{0.7501} & \underline{0.7973} & \underline{0.8646} & \underline{0.8986} & 0.9164 & 0.9315 & 0.9393 \\
    & \textbf{Concat-text-tab} & 0.6605 & 0.7495 & 0.7939 & 0.8526 & 0.8951 & \underline{0.922} & \underline{0.9381} & \underline{0.9501} \\
    \cline{2-10}
    & \textbf{C-BN-text} & 0.66 & 0.7472 & 0.7999 & *0.8705 & *0.9017 & 0.9199 & 0.9345 & 0.9434 \\
    & \textbf{V-BN-text} & 0.6521 & 0.7545 & *0.8091 & \textbf{*0.8829} & \textbf{*0.9141} & \textbf{*0.931} & \textbf{*0.9475} & \textbf{*0.9562} \\
    & \textbf{V-C-BN-text} & 0.6644 & \textbf{*0.7595} & \textbf{*0.8107} & *0.8802 & *0.9101 & *0.9267 & *0.9421 & *0.9515 \\
    \cline{2-10}
    & \textbf{change vs. baseline} & -2.61\% & +0.94\% & +1.35\% & +1.82\% & +1.54\% & +0.9\% & +0.94\% & +0.61\% \\
    \midrule
    \texttt{mean} & \textbf{BN-only} & 0.5904 & 0.5996 & 0.6111 & 0.6182 & 0.6228 & 0.6239 & 0.6247 & 0.6252 \\
    & \textbf{text-only} & \textbf{\underline{0.8173}} & \underline{0.8654} & \underline{0.8885} & \underline{0.915} & \underline{0.9298} & 0.9389 & 0.9456 & 0.9517 \\
    & \textbf{Concat-text-tab} & 0.7827 & 0.848 & 0.8828 & 0.9095 & 0.9289 & \underline{0.942} & \underline{0.9503} & \underline{0.9566} \\
    \cline{2-10}
    & \textbf{C-BN-text} & 0.8129 & *0.8666 & *0.8917 & *0.9194 & *0.933 & 0.941 & 0.9477 & 0.9538 \\
    & \textbf{V-BN-text} & 0.7974 & 0.8601 & 0.8867 & \textbf{*0.9235} & \textbf{*0.9398} & \textbf{*0.9488} & \textbf{*0.9558} & \textbf{*0.9615} \\
    & \textbf{V-C-BN-text} & 0.8131 & \textbf{*0.8707} & \textbf{*0.8945} & *0.923 & *0.9371 & *0.9458 & *0.9525 & *0.9586 \\
    \cline{2-10}
    & \textbf{change vs. baseline} & -0.42\% & +0.53\% & +0.6\% & +0.85\% & +1.0\% & +0.68\% & +0.55\% & +0.49\% \\
    \bottomrule
    \bottomrule
\end{tabular}}
\end{table*}

\begin{table*}[t]
  \caption{Brier scores ($\downdownarrows$) for the predictions of our models over $\mathcal{X}_{test}$, averaged over 20 seeds for various training sizes $n$. The best model per training size and per symptom is highlighted in \textbf{bold}. The best baseline model for each class is \underline{underlined}. Cases where a model outperforms the best baseline model significantly are indicated by \textbf{*} ($p < 0.05$ in a one-sided Wilcoxon signed-rank test over 20 seeds).}
  \label{tab:comparison_models_Brier_all}
\resizebox{\textwidth}{!}{
\begin{tabular}{llcccccccc}
    \toprule
    & & \multicolumn{8}{c}{Training size $n$} \\ \cmidrule{3-10}
    & & \textbf{100} & \textbf{187} & \textbf{350} & \textbf{654} & \textbf{1223} & \textbf{2287} & \textbf{4278} & \textbf{8000}\\
    \midrule
    \texttt{dysp} & \textbf{BN-only} & 0.0861 & 0.083 & 0.0797 & 0.0777 & 0.077 & 0.0766 & 0.0765 & 0.0763 \\
    & \textbf{text-only} & \underline{0.0485} & \underline{0.0365} & \underline{0.0314} & \underline{0.0277} & \underline{0.023} & \underline{0.019} & \underline{0.0172} & \underline{0.0128} \\
    & \textbf{Concat-text-tab} & 0.0508 & 0.0396 & 0.0339 & 0.0288 & 0.0251 & 0.0196 & 0.0172 & 0.0145 \\
    \cline{2-10}
    & \textbf{C-BN-text} & 0.0517 & 0.0367 & 0.0324 & 0.0275 & 0.0231 & 0.0192 & 0.0175 & 0.0132 \\
    & \textbf{V-BN-text} & 0.0498 & 0.0375 & 0.0333 & 0.0289 & 0.0249 & 0.02 & 0.0181 & 0.0145 \\
    & \textbf{V-C-BN-text} & \textbf{*0.0468} & \textbf{*0.0345} & \textbf{*0.0301} & \textbf{*0.0255} & \textbf{*0.0218} & \textbf{*0.0177} & \textbf{*0.0156} & \textbf{0.0123} \\
    \cline{2-10}
    & \textbf{change vs. baseline} & -0.16\% & -0.21\% & -0.12\% & -0.22\% & -0.12\% & -0.13\% & -0.15\% & -0.04\% \\
    \midrule
    \texttt{cough} & \textbf{BN-only} & 0.1267 & 0.1209 & 0.1176 & 0.1169 & 0.1158 & 0.1155 & 0.1155 & 0.1154 \\
    & \textbf{text-only} & \underline{0.0882} & \underline{0.0659} & 0.0576 & 0.0494 & 0.0401 & 0.0337 & 0.027 & 0.0247 \\
    & \textbf{Concat-text-tab} & 0.0907 & 0.0659 & \underline{0.0553} & \underline{0.0462} & \underline{0.0398} & \underline{0.0327} & \underline{0.0264} & \underline{0.0237} \\
    \cline{2-10}
    & \textbf{C-BN-text} & *0.0857 & *0.0642 & 0.0572 & 0.0487 & 0.0389 & 0.0328 & 0.0268 & 0.0244 \\
    & \textbf{V-BN-text} & *0.0831 & *0.0601 & *0.0529 & *0.044 & *0.0349 & *0.0296 & *0.0242 & 0.0229 \\
    & \textbf{V-C-BN-text} & \textbf{*0.0779} & \textbf{*0.0575} & \textbf{*0.0503} & \textbf{*0.0417} & \textbf{*0.0327} & \textbf{*0.0274} & \textbf{*0.0228} & \textbf{*0.0203} \\
    \cline{2-10}
    & \textbf{change vs. baseline} & -1.04\% & -0.84\% & -0.5\% & -0.45\% & -0.71\% & -0.54\% & -0.36\% & -0.34\% \\
    \midrule
    \texttt{pain} & \textbf{BN-only} & 0.1138 & 0.1119 & 0.1091 & 0.1078 & 0.1074 & 0.1073 & 0.1071 & 0.1072 \\
    & \textbf{text-only} & \textbf{\underline{0.0854}} & \underline{0.0711} & \textbf{\underline{0.0628}} & \underline{0.0553} & \underline{0.0491} & \underline{0.0451} & \underline{0.0426} & \underline{0.0383} \\
    & \textbf{Concat-text-tab} & 0.0942 & 0.0789 & 0.0666 & 0.0623 & 0.0524 & 0.0478 & 0.0436 & 0.0393 \\
    \cline{2-10}
    & \textbf{C-BN-text} & 0.0915 & 0.0734 & 0.0702 & 0.0604 & 0.0535 & 0.0501 & 0.047 & 0.0423 \\
    & \textbf{V-BN-text} & 0.0963 & 0.0792 & 0.0717 & 0.0577 & 0.0511 & 0.046 & 0.0427 & 0.0385 \\
    & \textbf{V-C-BN-text} & 0.0866 & \textbf{*0.0704} & 0.0647 & \textbf{0.0539} & \textbf{0.0483} & \textbf{0.0444} & \textbf{0.0414} & \textbf{0.0377} \\
    \cline{2-10}
    & \textbf{change vs. baseline} & +0.13\% & -0.07\% & +0.19\% & -0.14\% & -0.08\% & -0.06\% & -0.13\% & -0.06\% \\
    \midrule
    \texttt{nasal} & \textbf{BN-only} & 0.1216 & 0.1197 & 0.1185 & 0.1183 & 0.1175 & 0.1177 & 0.1174 & 0.1173 \\
    & \textbf{text-only} & \underline{0.0424} & \underline{0.0373} & \underline{0.0315} & \underline{0.0274} & \underline{0.0238} & \underline{0.0221} & \underline{0.019} & 0.0181 \\
    & \textbf{Concat-text-tab} & 0.0513 & 0.0417 & 0.035 & 0.0309 & 0.0262 & 0.0224 & 0.02 & \underline{0.0177} \\
    \cline{2-10}
    & \textbf{C-BN-text} & 0.0441 & *0.037 & *0.0311 & *0.0267 & *0.0235 & *0.0218 & 0.0189 & 0.018 \\
    & \textbf{V-BN-text} & 0.045 & *0.0344 & *0.0289 & 0.0262 & *0.0224 & *0.0201 & *0.018 & *0.0165 \\
    & \textbf{V-C-BN-text} & \textbf{*0.0395} & \textbf{*0.0336} & \textbf{*0.0279} & \textbf{*0.0244} & \textbf{*0.0214} & \textbf{*0.019} & \textbf{*0.0173} & \textbf{*0.0164} \\
    \cline{2-10}
    & \textbf{change vs. baseline} & -0.3\% & -0.37\% & -0.36\% & -0.3\% & -0.24\% & -0.31\% & -0.17\% & -0.13\% \\
    \midrule
    \texttt{fever} & \textbf{BN-only} & 0.3253 & 0.3153 & 0.3083 & 0.3049 & 0.3029 & 0.3016 & 0.301 & 0.3008 \\
    & \textbf{text-only} & \underline{0.256} & 0.2257 & 0.201 & 0.1744 & \underline{0.1448} & 0.1256 & 0.1126 & 0.0984 \\
    & \textbf{Concat-text-tab} & 0.2632 & \underline{0.2208} & \underline{0.1978} & \underline{0.1738} & 0.1449 & \underline{0.1243} & \underline{0.1081} & \underline{0.0962} \\
    \cline{2-10}
    & \textbf{C-BN-text} & 0.2683 & 0.2253 & 0.2089 & 0.1709 & 0.1441 & 0.1261 & 0.1144 & 0.1038 \\
    & \textbf{V-BN-text} & 0.2868 & 0.2327 & 0.2098 & *0.1613 & *0.1353 & 0.1242 & 0.1067 & 0.0986 \\
    & \textbf{V-C-BN-text} & \textbf{0.2535} & \textbf{*0.2115} & \textbf{*0.1911} & \textbf{*0.1511} & \textbf{*0.1269} & \textbf{*0.1146} & \textbf{*0.1014} & \textbf{0.0941} \\
    \cline{2-10}
    & \textbf{change vs. baseline} & -0.25\% & -0.93\% & -0.67\% & -2.27\% & -1.79\% & -0.96\% & -0.68\% & -0.22\% \\
    \midrule
    \texttt{mean} & \textbf{BN-only} & 0.1547 & 0.1502 & 0.1466 & 0.1451 & 0.1441 & 0.1437 & 0.1435 & 0.1434 \\
    & \textbf{text-only} & \underline{0.1041} & \underline{0.0873} & \underline{0.0768} & \underline{0.0668} & \underline{0.0562} & \underline{0.0491} & 0.0437 & 0.0384 \\
    & \textbf{Concat-text-tab} & 0.11 & 0.0894 & 0.0777 & 0.0684 & 0.0577 & 0.0494 & \underline{0.0431} & \underline{0.0383} \\
    \cline{2-10}
    & \textbf{C-BN-text} & 0.1082 & 0.0873 & 0.08 & 0.0668 & 0.0566 & 0.05 & 0.0449 & 0.0404 \\
    & \textbf{V-BN-text} & 0.1122 & 0.0888 & 0.0793 & *0.0636 & *0.0537 & 0.048 & *0.0419 & 0.0382 \\
    & \textbf{V-C-BN-text} & \textbf{*0.1009} & \textbf{*0.0815} & \textbf{*0.0728} & \textbf{*0.0593} & \textbf{*0.0502} & \textbf{*0.0446} & \textbf{*0.0397} & \textbf{*0.0362} \\
    \cline{2-10}
    & \textbf{change vs. baseline} & -0.32\% & -0.58\% & -0.4\% & -0.75\% & -0.6\% & -0.45\% & -0.34\% & -0.21\% \\
    \bottomrule
    \bottomrule
\end{tabular}}
\end{table*}

\subsection{Ground truth SimSUM Bayesian network} \label{sec:app_GT_BN}

The tabular portion of the SimSUM dataset was generated using a fully expert-defined Bayesian network, where not only the relations between the variables were defined by an expert, but also the conditional probabilities. In a real setting, it would be unrealistic to assume that we have access to this true generating process, which is why we learn the distributions from the data in our work, which gives us the \textbf{BN-only} model. However, having access to the true ground truth distributions allows us to swap the \textbf{BN-only} model for the \textbf{GT-only} model (GT indicating ground truth), and use its predictions to build the fusion models \textbf{C-GT-text}, \textbf{V-GT-text}, and \textbf{V-C-GT-text}. We do not change anything about the \textbf{text-only} model. Table \ref{tab:comparison_models_average_precision_gtbn} and Table \ref{tab:comparison_models_brier_gtbn} show the results. We note that these fusion models with access to the ground truth BN \textbf{GT-only} often outperform the models using \textbf{BN-only} as the Bayesian network, especially for smaller training sizes, as the BN better matches the true data generating process.

\begin{table*}[t]
  \caption{Average precision ($\upuparrows$) for the predictions of our models using the ground truth Bayesian network (\textbf{GT-only}) over $\mathcal{X}_{test}$, averaged over 20 seeds for various training sizes $n$. The best model per training size and per symptom is highlighted in \textbf{bold}. The best baseline model for each class is \underline{underlined}. Cases where a model outperforms the best baseline model significantly are indicated by \textbf{*} ($p < 0.05$ in a one-sided Wilcoxon signed-rank test over 20 seeds).}
  \label{tab:comparison_models_average_precision_gtbn}
\resizebox{\textwidth}{!}{
\begin{tabular}{llcccccccc}
    \toprule
    & & \multicolumn{8}{c}{Training size $n$} \\ \cmidrule{3-10}
    & & \textbf{100} & \textbf{187} & \textbf{350} & \textbf{654} & \textbf{1223} & \textbf{2287} & \textbf{4278} & \textbf{8000}\\
    \midrule
    \texttt{dysp} & \textbf{GT-only} & 0.7996 & 0.7996 & 0.7996 & 0.7996 & 0.7996 & 0.7996 & 0.7996 & 0.7996 \\
    & \textbf{text-only} & \underline{0.9246} & \underline{0.9482} & \underline{0.9578} & \underline{0.967} & 0.9731 & 0.9798 & 0.9831 & \underline{0.9878} \\
    & \textbf{Concat-text-tab} & 0.9127 & 0.9398 & 0.9533 & 0.9657 & \underline{0.9737} & \underline{0.9801} & \underline{0.9841} & 0.987 \\
    \cline{2-10}
    & \textbf{C-GT-text} & *0.9292 & *0.9499 & *0.9584 & 0.9676 & 0.974 & 0.9786 & 0.982 & 0.9875 \\
    & \textbf{V-GT-text} & 0.9283 & *0.9514 & 0.9572 & 0.9677 & 0.9742 & 0.9805 & 0.984 & *0.9881 \\
    & \textbf{V-C-GT-text} & \textbf{*0.9313} & \textbf{*0.9533} & \textbf{*0.9606} & \textbf{*0.97} & \textbf{*0.9757} & \textbf{*0.9817} & \textbf{*0.9851} & \textbf{*0.989} \\
    \cline{2-10}
    & \textbf{change vs. baseline} & +0.67\% & +0.5\% & +0.29\% & +0.3\% & +0.2\% & +0.16\% & +0.1\% & +0.12\% \\
    \midrule
    \texttt{cough} & \textbf{GT-only} & 0.797 & 0.797 & 0.797 & 0.797 & 0.797 & 0.797 & 0.797 & 0.797 \\
    & \textbf{text-only} & \underline{0.902} & \underline{0.9452} & 0.9571 & 0.9689 & 0.978 & 0.9826 & 0.9863 & 0.989 \\
    & \textbf{Concat-text-tab} & 0.8866 & 0.9393 & \underline{0.9582} & \underline{0.9705} & \underline{0.9794} & \underline{0.9851} & \underline{0.9886} & \underline{0.9908} \\
    \cline{2-10}
    & \textbf{C-GT-text} & *0.9175 & *0.9501 & *0.9612 & 0.9704 & 0.9776 & 0.9816 & 0.986 & 0.9882 \\
    & \textbf{V-GT-text} & \textbf{*0.9279} & \textbf{*0.9575} & *0.9658 & \textbf{*0.9754} & \textbf{*0.9831} & \textbf{*0.9868} & \textbf{*0.9898} & \textbf{*0.9919} \\
    & \textbf{V-C-GT-text} & *0.9269 & *0.9571 & \textbf{*0.9663} & *0.9752 & *0.9823 & *0.9859 & 0.9891 & *0.9914 \\
    \cline{2-10}
    & \textbf{change vs. baseline} & +2.59\% & +1.24\% & +0.81\% & +0.49\% & +0.37\% & +0.17\% & +0.12\% & +0.11\% \\
    \midrule
    \texttt{pain} & \textbf{GT-only} & 0.3603 & 0.3603 & 0.3603 & 0.3603 & 0.3603 & 0.3603 & 0.3603 & 0.3603 \\
    & \textbf{text-only} & \underline{0.6181} & \underline{0.7252} & \underline{0.7648} & \underline{0.804} & \underline{0.825} & 0.8377 & 0.8471 & 0.8608 \\
    & \textbf{Concat-text-tab} & 0.5091 & 0.6537 & 0.7422 & 0.7858 & 0.8183 & \underline{0.8412} & \underline{0.8559} & \underline{0.868} \\
    \cline{2-10}
    & \textbf{C-GT-text} & *0.6259 & \textbf{*0.7328} & \textbf{*0.7739} & *0.8121 & *0.8322 & 0.8431 & 0.852 & 0.8651 \\
    & \textbf{V-GT-text} & 0.6119 & *0.7308 & 0.7628 & \textbf{*0.8215} & \textbf{*0.8473} & \textbf{*0.8601} & \textbf{*0.8699} & \textbf{*0.8825} \\
    & \textbf{V-C-GT-text} & \textbf{*0.6296} & *0.7326 & *0.7708 & *0.8149 & *0.8375 & *0.8508 & *0.8602 & *0.8734 \\
    \cline{2-10}
    & \textbf{change vs. baseline} & +1.15\% & +0.75\% & +0.92\% & +1.76\% & +2.23\% & +1.89\% & +1.4\% & +1.45\% \\
    \midrule
    \texttt{nasal} & \textbf{GT-only} & 0.6477 & 0.6477 & 0.6477 & 0.6477 & 0.6477 & 0.6477 & 0.6477 & 0.6477 \\
    & \textbf{text-only} & \underline{0.9512} & \underline{0.9584} & 0.9657 & 0.9706 & 0.9743 & 0.9779 & 0.98 & 0.9816 \\
    & \textbf{Concat-text-tab} & 0.9444 & 0.9579 & \underline{0.9666} & \underline{0.973} & \underline{0.9777} & \underline{0.9818} & \underline{0.9847} & \underline{0.9869} \\
    \cline{2-10}
    & \textbf{C-GT-text} & \textbf{*0.9559} & *0.9625 & 0.9689 & *0.9771 & *0.9799 & 0.9817 & 0.9842 & 0.9847 \\
    & \textbf{V-GT-text} & *0.9538 & \textbf{*0.9644} & \textbf{*0.9715} & \textbf{*0.9785} & \textbf{*0.983} & \textbf{*0.9861} & \textbf{*0.9879} & \textbf{*0.9885} \\
    & \textbf{V-C-GT-text} & *0.9548 & *0.9632 & *0.9703 & *0.9758 & *0.9799 & *0.9836 & *0.9857 & 0.987 \\
    \cline{2-10}
    & \textbf{change vs. baseline} & +0.47\% & +0.6\% & +0.49\% & +0.55\% & +0.53\% & +0.43\% & +0.31\% & +0.16\% \\
    \midrule
    \texttt{fever} & \textbf{GT-only} & 0.5465 & 0.5465 & 0.5465 & 0.5465 & 0.5465 & 0.5465 & 0.5465 & 0.5465 \\
    & \textbf{text-only} & \underline{0.6905} & \underline{0.7501} & \underline{0.7973} & \underline{0.8646} & \underline{0.8986} & 0.9164 & 0.9315 & 0.9393 \\
    & \textbf{Concat-text-tab} & 0.6605 & 0.7495 & 0.7939 & 0.8526 & 0.8951 & \underline{0.922} & \underline{0.9381} & \underline{0.9501} \\
    \cline{2-10}
    & \textbf{C-GT-text} & 0.681 & 0.7536 & *0.803 & *0.8711 & *0.9018 & 0.9199 & 0.9344 & 0.9434 \\
    & \textbf{V-GT-text} & \textbf{*0.7149} & \textbf{*0.7869} & \textbf{*0.8224} & \textbf{*0.8861} & \textbf{*0.9154} & \textbf{*0.9313} & \textbf{*0.9476} & \textbf{*0.9563} \\
    & \textbf{V-C-GT-text} & 0.6936 & *0.7756 & *0.8179 & *0.8814 & *0.9103 & *0.9266 & *0.942 & *0.9515 \\
    \cline{2-10}
    & \textbf{change vs. baseline} & +2.44\% & +3.68\% & +2.52\% & +2.15\% & +1.68\% & +0.93\% & +0.95\% & +0.62\% \\
    \midrule
    \texttt{mean} & \textbf{GT-only} & 0.6302 & 0.6302 & 0.6302 & 0.6302 & 0.6302 & 0.6302 & 0.6302 & 0.6302 \\
    & \textbf{text-only} & \underline{0.8173} & \underline{0.8654} & \underline{0.8885} & \underline{0.915} & \underline{0.9298} & 0.9389 & 0.9456 & 0.9517 \\
    & \textbf{Concat-text-tab} & 0.7827 & 0.848 & 0.8828 & 0.9095 & 0.9289 & \underline{0.942} & \underline{0.9503} & \underline{0.9566} \\
    \cline{2-10}
    & \textbf{C-GT-text} & *0.8219 & *0.8698 & *0.8931 & *0.9197 & *0.9331 & 0.941 & 0.9477 & 0.9538 \\
    & \textbf{V-GT-text} & \textbf{*0.8273} & \textbf{*0.8782} & *0.8959 & \textbf{*0.9258} & \textbf{*0.9406} & \textbf{*0.949} & \textbf{*0.9558} & \textbf{*0.9615} \\
    & \textbf{V-C-GT-text} & *0.8272 & *0.8764 & \textbf{*0.8972} & *0.9234 & *0.9372 & *0.9457 & *0.9524 & *0.9585 \\
    \cline{2-10}
    & \textbf{change vs. baseline} & +1.01\% & +1.28\% & +0.87\% & +1.08\% & +1.08\% & +0.69\% & +0.55\% & +0.49\% \\
    \bottomrule
    \bottomrule
\end{tabular}}
\end{table*}

\begin{table*}[t]
  \caption{Brier scores ($\downdownarrows$) for the predictions of our models using the ground truth Bayesian network (\textbf{GT-only}) over $\mathcal{X}_{test}$, averaged over 20 seeds for various training sizes $n$. The best model per training size and per symptom is highlighted in \textbf{bold}. The best baseline model for each class is \underline{underlined}. Cases where a model outperforms the best baseline model significantly are indicated by \textbf{*} ($p < 0.05$ in a one-sided Wilcoxon signed-rank test over 20 seeds).}
  \label{tab:comparison_models_brier_gtbn}
\resizebox{\textwidth}{!}{
\begin{tabular}{llcccccccc}
    \toprule
    & & \multicolumn{8}{c}{Training size $n$} \\ \cmidrule{3-10}
    & & \textbf{100} & \textbf{187} & \textbf{350} & \textbf{654} & \textbf{1223} & \textbf{2287} & \textbf{4278} & \textbf{8000}\\
    \midrule
    \texttt{dysp} & \textbf{GT-only} & 0.0761 & 0.0761 & 0.0761 & 0.0761 & 0.0761 & 0.0761 & 0.0761 & 0.0761 \\
    & \textbf{text-only} & \underline{0.0485} & \underline{0.0365} & \underline{0.0314} & \underline{0.0277} & \underline{0.023} & \underline{0.019} & \underline{0.0172} & \underline{0.0128} \\
    & \textbf{Concat-text-tab} & 0.0508 & 0.0396 & 0.0339 & 0.0288 & 0.0251 & 0.0196 & 0.0172 & 0.0145 \\
    \cline{2-10}
    & \textbf{C-GT-text} & 0.0508 & 0.0365 & 0.0322 & 0.0274 & 0.023 & 0.0192 & 0.0175 & 0.0132 \\
    & \textbf{V-GT-text} & 0.0468 & *0.0352 & 0.0324 & 0.0287 & 0.0248 & 0.02 & 0.0183 & 0.0148 \\
    & \textbf{V-C-GT-text} & \textbf{*0.0458} & \textbf{*0.0337} & \textbf{*0.0298} & \textbf{*0.0254} & \textbf{*0.0218} & \textbf{*0.0178} & \textbf{*0.0157} & \textbf{0.0124} \\
    \cline{2-10}
    & \textbf{change vs. baseline} & -0.27\% & -0.29\% & -0.16\% & -0.23\% & -0.12\% & -0.13\% & -0.15\% & -0.03\% \\
    \midrule
    \texttt{cough} & \textbf{GT-only} & 0.1146 & 0.1146 & 0.1146 & 0.1146 & 0.1146 & 0.1146 & 0.1146 & 0.1146 \\
    & \textbf{text-only} & \underline{0.0882} & \underline{0.0659} & 0.0576 & 0.0494 & 0.0401 & 0.0337 & 0.027 & 0.0247 \\
    & \textbf{Concat-text-tab} & 0.0907 & 0.0659 & \underline{0.0553} & \underline{0.0462} & \underline{0.0398} & \underline{0.0327} & \underline{0.0264} & \underline{0.0237} \\
    \cline{2-10}
    & \textbf{C-GT-text} & *0.085 & *0.064 & 0.0571 & 0.0486 & 0.0388 & 0.0328 & 0.0268 & 0.0244 \\
    & \textbf{V-GT-text} & \textbf{*0.0727} & \textbf{*0.0555} & *0.0496 & *0.042 & *0.0341 & *0.0292 & *0.0239 & 0.0224 \\
    & \textbf{V-C-GT-text} & *0.0742 & *0.0558 & \textbf{*0.049} & \textbf{*0.0408} & \textbf{*0.0324} & \textbf{*0.0272} & \textbf{*0.0228} & \textbf{*0.0202} \\
    \cline{2-10}
    & \textbf{change vs. baseline} & -1.55\% & -1.04\% & -0.62\% & -0.54\% & -0.74\% & -0.55\% & -0.37\% & -0.35\% \\
    \midrule
    \texttt{pain} & \textbf{GT-only} & 0.1066 & 0.1066 & 0.1066 & 0.1066 & 0.1066 & 0.1066 & 0.1066 & 0.1066 \\
    & \textbf{text-only} & \textbf{\underline{0.0854}} & \underline{0.0711} & \textbf{\underline{0.0628}} & \underline{0.0553} & \underline{0.0491} & \underline{0.0451} & \underline{0.0426} & \underline{0.0383} \\
    & \textbf{Concat-text-tab} & 0.0942 & 0.0789 & 0.0666 & 0.0623 & 0.0524 & 0.0478 & 0.0436 & 0.0393 \\
    \cline{2-10}
    & \textbf{C-GT-text} & 0.0919 & 0.0736 & 0.0702 & 0.0604 & 0.0536 & 0.0501 & 0.047 & 0.0423 \\
    & \textbf{V-GT-text} & 0.0952 & 0.0767 & 0.0705 & 0.0573 & 0.0508 & 0.0457 & 0.0427 & 0.0384 \\
    & \textbf{V-C-GT-text} & 0.0866 & \textbf{*0.0699} & 0.0644 & \textbf{0.0538} & \textbf{0.0482} & \textbf{0.0444} & \textbf{0.0414} & \textbf{*0.0377} \\
    \cline{2-10}
    & \textbf{change vs. baseline} & +0.13\% & -0.12\% & +0.16\% & -0.14\% & -0.09\% & -0.07\% & -0.12\% & -0.07\% \\
    \midrule
    \texttt{nasal} & \textbf{GT-only} & 0.1167 & 0.1167 & 0.1167 & 0.1167 & 0.1167 & 0.1167 & 0.1167 & 0.1167 \\
    & \textbf{text-only} & \underline{0.0424} & \underline{0.0373} & \underline{0.0315} & \underline{0.0274} & \underline{0.0238} & \underline{0.0221} & \underline{0.019} & 0.0181 \\
    & \textbf{Concat-text-tab} & 0.0513 & 0.0417 & 0.035 & 0.0309 & 0.0262 & 0.0224 & 0.02 & \underline{0.0177} \\
    \cline{2-10}
    & \textbf{C-GT-text} & 0.0441 & *0.037 & *0.0312 & *0.0268 & *0.0235 & *0.0218 & 0.0189 & 0.018 \\
    & \textbf{V-GT-text} & 0.0417 & \textbf{*0.0331} & *0.0281 & *0.0255 & *0.0222 & *0.0201 & *0.0179 & *0.0164 \\
    & \textbf{V-C-GT-text} & \textbf{*0.039} & *0.0335 & \textbf{*0.0278} & \textbf{*0.0245} & \textbf{*0.0214} & \textbf{*0.019} & \textbf{*0.0173} & \textbf{*0.0164} \\
    \cline{2-10}
    & \textbf{change vs. baseline} & -0.35\% & -0.42\% & -0.37\% & -0.29\% & -0.25\% & -0.31\% & -0.17\% & -0.14\% \\
    \midrule
    \texttt{fever} & \textbf{GT-only} & 0.3006 & 0.3006 & 0.3006 & 0.3006 & 0.3006 & 0.3006 & 0.3006 & 0.3006 \\
    & \textbf{text-only} & \underline{0.256} & 0.2257 & 0.201 & 0.1744 & \underline{0.1448} & 0.1256 & 0.1126 & 0.0984 \\
    & \textbf{Concat-text-tab} & 0.2632 & \underline{0.2208} & \underline{0.1978} & \underline{0.1738} & 0.1449 & \underline{0.1243} & \underline{0.1081} & \underline{0.0962} \\
    \cline{2-10}
    & \textbf{C-GT-text} & 0.268 & 0.2253 & 0.2088 & 0.1709 & 0.1441 & 0.1261 & 0.1144 & 0.1038 \\
    & \textbf{V-GT-text} & 0.2766 & 0.2241 & 0.2054 & *0.1589 & *0.1338 & 0.124 & 0.1065 & 0.0985 \\
    & \textbf{V-C-GT-text} & \textbf{0.2503} & \textbf{*0.2092} & \textbf{*0.1897} & \textbf{*0.1505} & \textbf{*0.1266} & \textbf{*0.1148} & \textbf{*0.1015} & \textbf{0.0942} \\
    \cline{2-10}
    & \textbf{change vs. baseline} & -0.58\% & -1.16\% & -0.81\% & -2.34\% & -1.81\% & -0.95\% & -0.67\% & -0.21\% \\
    \midrule
    \texttt{mean} & \textbf{GT-only} & 0.1429 & 0.1429 & 0.1429 & 0.1429 & 0.1429 & 0.1429 & 0.1429 & 0.1429 \\
    & \textbf{text-only} & \underline{0.1041} & \underline{0.0873} & \underline{0.0768} & \underline{0.0668} & \underline{0.0562} & \underline{0.0491} & 0.0437 & 0.0384 \\
    & \textbf{Concat-text-tab} & 0.11 & 0.0894 & 0.0777 & 0.0684 & 0.0577 & 0.0494 & \underline{0.0431} & \underline{0.0383} \\
    \cline{2-10}
    & \textbf{C-GT-text} & 0.108 & 0.0873 & 0.0799 & 0.0668 & 0.0566 & 0.05 & 0.0449 & 0.0404 \\
    & \textbf{V-GT-text} & 0.1066 & *0.0849 & 0.0772 & *0.0625 & *0.0531 & *0.0478 & *0.0419 & 0.0381 \\
    & \textbf{V-C-GT-text} & \textbf{*0.0992} & \textbf{*0.0804} & \textbf{*0.0721} & \textbf{*0.059} & \textbf{*0.0501} & \textbf{*0.0446} & \textbf{*0.0397} & \textbf{*0.0362} \\
    \cline{2-10}
    & \textbf{change vs. baseline} & -0.49\% & -0.69\% & -0.47\% & -0.78\% & -0.61\% & -0.45\% & -0.33\% & -0.21\% \\
    \bottomrule
    \bottomrule
\end{tabular}}
\end{table*}

\subsection{Present vs. mentioned subsets}\label{sec:app_PM_subsets}

We report the Brier scores of the \textbf{text-only} baseline and our BN + text models on the subsets \{\textit{present, mentioned}; \textit{present, not mentioned}; \textit{not present, mentioned}; \textit{not present, not mentioned}\} in Table \ref{tab:comparison_models_brier_pm}, Table \ref{tab:comparison_models_brier_pnm}, Table \ref{tab:comparison_models_brier_npm}, and Table \ref{tab:comparison_models_brier_npnm} respectively.

As mentioned in Section \ref{sec:pvm_subsets}, \textbf{V-C-BN-text} improves significantly over the \textbf{text-only} classifiers in the \textit{present, not mentioned} subset for small training sizes, but this breaks down at larger training sizes for \texttt{pain}, \texttt{nasal}, and \texttt{fever}. We attribute this to two factors: (1) the BN does not increase in accuracy as much as the \textbf{text-only} classifiers with more training data, leading the consistency node to favor the predictions of the neural classifiers, and (2) the \textbf{text-only} classifiers become more confident at higher training sizes, leading their contributions as virtual evidence to be weighed more heavily.

Evidence of (1) can be seen in Table \ref{tab:comparison_models_average_precision_all}, where the performance of the BN does not improve as much as the \textbf{text-only} classifiers at higher training sizes. Note this table also helps indicate why \texttt{dysp} and \texttt{cough} continue to display larger improvements on the \textit{present, not mentioned} than the other symptoms: the BN is better at predicting those symptoms relative to the neural classifiers at higher training sizes.

As evidence of (2), we define a standard confidence measure using normalized entropy \citep{wu2021using}: using standard Shannon entropy to determine the uncertainty of our model's predictions:

\begin{small}
\begin{align}
    H(\mathbf{p}) = - \sum_{i=1}^{K} p_i \log p_i
\label{eq:entropy}
\end{align}
\end{small}

we define confidence as 1 minus the normalized entropy:

\begin{small}
\begin{align}
    1 - \frac{H(\mathbf{p})}{\log K}
\label{eq:confidence}
\end{align}
\end{small}

We report the confidence of the \textbf{text-only} classifiers (substituting $\mathcal{P}(T_{s_j} \mid note)$ for $p$ in Equation \ref{eq:confidence}) in Table \ref{tab:comparison_models_confidence}, which shows that the neural classifiers become more confident at larger training sizes. As noted in Section \ref{sec:pvm_subsets}, confident virtual evidence can lead to a more confident final prediction, producing a more confidently wrong prediction in the case of faulty virtual evidence. As shown in Table \ref{tab:comparison_models_brier_pnm}, the consistency node can help correct for this weakness of virtual evidence at higher training sizes for the more difficult symptoms \texttt{Pain} and \texttt{Fever}.

\begin{table*}[t]
  \caption{Brier scores ($\downdownarrows$) for our models on the \textit{present, mentioned} subset across various training sizes. The best model per training size and per symptom is highlighted in \textbf{bold}. Cases where a model outperforms \textbf{text-only} significantly are indicated by \textbf{*} ($p < 0.05$ in a one-sided Wilcoxon signed-rank test over 20 seeds).}
  \label{tab:comparison_models_brier_pm}
\resizebox{\textwidth}{!}{
\begin{tabular}{llcccccccc}
    \toprule
    & & \multicolumn{8}{c}{Training size $n$} \\ \cmidrule{3-10}
    & & \textbf{100} & \textbf{187} & \textbf{350} & \textbf{654} & \textbf{1223} & \textbf{2287} & \textbf{4278} & \textbf{8000}\\
    \midrule
    \texttt{dysp} & \textbf{text-only} & 0.1174 & \textbf{0.08} & \textbf{0.0719} & 0.0648 & \textbf{0.0562} & \textbf{0.0369} & \textbf{0.0355} & \textbf{0.0283} \\
    & \textbf{C-BN-text} & 0.1411 & 0.0912 & 0.083 & 0.0695 & 0.0592 & 0.0431 & 0.0419 & 0.0309 \\
    & \textbf{V-BN-text} & 0.1477 & 0.1031 & 0.0931 & 0.0831 & 0.0719 & 0.0487 & 0.0465 & 0.0364 \\
    & \textbf{V-C-BN-text} & \textbf{0.115} & 0.0807 & 0.0742 & \textbf{0.064} & 0.0567 & 0.0433 & 0.04 & 0.0308 \\
    \cline{2-10}
    & \textbf{change vs. baseline} & -0.24\% & +0.07\% & +0.23\% & -0.08\% & +0.05\% & +0.62\% & +0.45\% & +0.25\% \\
    \midrule
    \texttt{cough} & \textbf{text-only} & 0.1153 & 0.0659 & 0.0592 & 0.0402 & 0.0283 & 0.0182 & 0.017 & 0.0098 \\
    & \textbf{C-BN-text} & 0.1259 & 0.0733 & 0.0634 & 0.0473 & 0.0328 & 0.0217 & 0.0183 & 0.0135 \\
    & \textbf{V-BN-text} & 0.1228 & 0.0668 & 0.0563 & \textbf{*0.0365} & \textbf{*0.0243} & \textbf{*0.0145} & \textbf{*0.0139} & \textbf{0.0088} \\
    & \textbf{V-C-BN-text} & \textbf{*0.1031} & \textbf{0.0627} & \textbf{*0.0521} & 0.0377 & 0.025 & 0.0161 & 0.0147 & 0.0098 \\
    \cline{2-10}
    & \textbf{change vs. baseline} & -1.22\% & -0.32\% & -0.71\% & -0.38\% & -0.4\% & -0.37\% & -0.31\% & -0.1\% \\
    \midrule
    \texttt{pain} & \textbf{text-only} & \textbf{0.3809} & \textbf{0.2508} & \textbf{0.1941} & \textbf{0.1393} & \textbf{0.1252} & \textbf{0.103} & \textbf{0.0909} & \textbf{0.0706} \\
    & \textbf{C-BN-text} & 0.4679 & 0.3097 & 0.3086 & 0.2337 & 0.1912 & 0.1696 & 0.1489 & 0.1162 \\
    & \textbf{V-BN-text} & 0.5569 & 0.401 & 0.3573 & 0.2387 & 0.1921 & 0.15 & 0.1247 & 0.0931 \\
    & \textbf{V-C-BN-text} & 0.4157 & 0.28 & 0.2625 & 0.1915 & 0.1608 & 0.1354 & 0.1147 & 0.0885 \\
    \cline{2-10}
    & \textbf{change vs. baseline} & +3.48\% & +2.91\% & +6.84\% & +5.22\% & +3.56\% & +3.24\% & +2.38\% & +1.79\% \\
    \midrule
    \texttt{nasal} & \textbf{text-only} & \textbf{0.0521} & \textbf{0.0349} & \textbf{0.0205} & 0.019 & \textbf{0.0124} & \textbf{0.0065} & \textbf{0.0057} & 0.004 \\
    & \textbf{C-BN-text} & 0.0681 & 0.041 & 0.0281 & 0.0222 & 0.0163 & 0.0112 & 0.0086 & 0.007 \\
    & \textbf{V-BN-text} & 0.0924 & 0.0501 & 0.0328 & 0.0287 & 0.0178 & 0.009 & 0.0067 & \textbf{0.0038} \\
    & \textbf{V-C-BN-text} & 0.054 & 0.037 & 0.0234 & \textbf{0.0189} & 0.014 & 0.0083 & 0.0067 & 0.0046 \\
    \cline{2-10}
    & \textbf{change vs. baseline} & +0.19\% & +0.21\% & +0.29\% & -0.01\% & +0.16\% & +0.18\% & +0.09\% & -0.03\% \\
    \midrule
    \texttt{fever} & \textbf{text-only} & \textbf{0.7589} & \textbf{0.5334} & \textbf{0.4449} & \textbf{0.2503} & \textbf{0.1718} & \textbf{0.155} & \textbf{0.0998} & \textbf{0.0722} \\
    & \textbf{C-BN-text} & 0.8994 & 0.6497 & 0.5861 & 0.3683 & 0.2582 & 0.2037 & 0.1579 & 0.1223 \\
    & \textbf{V-BN-text} & 1.0834 & 0.7634 & 0.6551 & 0.3741 & 0.2486 & 0.2152 & 0.1294 & 0.0905 \\
    & \textbf{V-C-BN-text} & 0.8151 & 0.5772 & 0.4998 & 0.2932 & 0.1965 & 0.1643 & 0.1076 & 0.0774 \\
    \cline{2-10}
    & \textbf{change vs. baseline} & +5.62\% & +4.38\% & +5.49\% & +4.29\% & +2.47\% & +0.93\% & +0.78\% & +0.52\% \\
    \midrule
    \texttt{mean} & \textbf{text-only} & \textbf{0.2849} & \textbf{0.193} & \textbf{0.1581} & \textbf{0.1027} & \textbf{0.0788} & \textbf{0.0639} & \textbf{0.0498} & \textbf{0.037} \\
    & \textbf{C-BN-text} & 0.3405 & 0.233 & 0.2139 & 0.1482 & 0.1115 & 0.0898 & 0.0751 & 0.058 \\
    & \textbf{V-BN-text} & 0.4007 & 0.2769 & 0.2389 & 0.1522 & 0.1109 & 0.0875 & 0.0642 & 0.0465 \\
    & \textbf{V-C-BN-text} & 0.3006 & 0.2075 & 0.1824 & 0.1211 & 0.0906 & 0.0735 & 0.0567 & 0.0422 \\
    \cline{2-10}
    & \textbf{change vs. baseline} & +1.57\% & +1.45\% & +2.43\% & +1.83\% & +1.18\% & +0.96\% & +0.69\% & +0.52\% \\
    \bottomrule
    \bottomrule
\end{tabular}}
\end{table*}

\begin{table*}[t]
  \caption{Brier scores ($\downdownarrows$) for our models on the \textit{present, not mentioned} subset across various training sizes. The best model per training size and per symptom is highlighted in \textbf{bold}. Cases where a model outperforms \textbf{text-only} significantly are indicated by \textbf{*} ($p < 0.05$ in a one-sided Wilcoxon signed-rank test over 20 seeds).}
  \label{tab:comparison_models_brier_pnm}
\resizebox{\textwidth}{!}{
\begin{tabular}{llcccccccc}
    \toprule
    & & \multicolumn{8}{c}{Training size $n$} \\ \cmidrule{3-10}
    & & \textbf{100} & \textbf{187} & \textbf{350} & \textbf{654} & \textbf{1223} & \textbf{2287} & \textbf{4278} & \textbf{8000}\\
    \midrule
    \texttt{dysp} & \textbf{text-only} & 0.8879 & 0.8893 & 0.8958 & 0.9192 & 0.8972 & 0.8517 & 0.8501 & 0.847 \\
    & \textbf{C-BN-text} & *0.7689 & *0.7967 & *0.7684 & *0.804 & *0.7996 & *0.7884 & *0.7733 & *0.7704 \\
    & \textbf{V-BN-text} & \textbf{*0.7117} & \textbf{*0.709} & \textbf{*0.7026} & \textbf{*0.7335} & \textbf{*0.7047} & \textbf{*0.6328} & \textbf{*0.6443} & \textbf{*0.6146} \\
    & \textbf{V-C-BN-text} & *0.7241 & *0.7441 & *0.7389 & *0.7826 & *0.7559 & *0.7284 & *0.7318 & *0.7038 \\
    \cline{2-10}
    & \textbf{change vs. baseline} & -17.62\% & -18.03\% & -19.31\% & -18.57\% & -19.25\% & -21.89\% & -20.58\% & -23.24\% \\
    \midrule
    \texttt{cough} & \textbf{text-only} & 0.5749 & 0.642 & 0.6555 & 0.644 & 0.66 & 0.6452 & 0.6715 & 0.6648 \\
    & \textbf{C-BN-text} & \textbf{*0.4496} & *0.5267 & \textbf{*0.5064} & \textbf{*0.535} & *0.56 & *0.561 & *0.5787 & *0.5909 \\
    & \textbf{V-BN-text} & *0.4654 & \textbf{*0.5249} & *0.5272 & *0.5376 & \textbf{*0.5587} & \textbf{*0.544} & \textbf{*0.5533} & \textbf{*0.5292} \\
    & \textbf{V-C-BN-text} & *0.4714 & *0.5556 & *0.5518 & *0.5751 & *0.5946 & *0.5921 & *0.6008 & *0.6111 \\
    \cline{2-10}
    & \textbf{change vs. baseline} & -12.53\% & -11.71\% & -14.91\% & -10.89\% & -10.13\% & -10.12\% & -11.81\% & -13.56\% \\
    \midrule
    \texttt{pain} & \textbf{text-only} & 0.7786 & 0.8574 & 0.8387 & 0.8591 & 0.8822 & 0.8734 & 0.8711 & 0.8674 \\
    & \textbf{C-BN-text} & *0.7251 & \textbf{*0.8035} & \textbf{*0.7825} & \textbf{*0.8102} & \textbf{*0.8236} & \textbf{*0.8236} & \textbf{*0.8206} & \textbf{*0.8178} \\
    & \textbf{V-BN-text} & 0.8285 & 0.8962 & 0.8896 & 0.9142 & 0.9304 & 0.922 & 0.9188 & 0.9092 \\
    & \textbf{V-C-BN-text} & \textbf{*0.724} & *0.8182 & *0.8038 & 0.8442 & *0.863 & 0.8695 & 0.8682 & 0.8679 \\
    \cline{2-10}
    & \textbf{change vs. baseline} & -5.46\% & -5.39\% & -5.62\% & -4.89\% & -5.86\% & -4.98\% & -5.05\% & -4.96\% \\
    \midrule
    \texttt{nasal} & \textbf{text-only} & 0.8839 & 0.9129 & 0.9198 & 0.9324 & 0.9298 & 0.899 & 0.9087 & 0.8948 \\
    & \textbf{C-BN-text} & \textbf{*0.7955} & \textbf{*0.8495} & \textbf{*0.8482} & \textbf{*0.8553} & \textbf{*0.8609} & *0.8463 & *0.8503 & *0.8434 \\
    & \textbf{V-BN-text} & *0.8381 & *0.8655 & *0.8625 & *0.8741 & *0.8785 & \textbf{*0.8297} & \textbf{*0.8262} & \textbf{*0.809} \\
    & \textbf{V-C-BN-text} & *0.8217 & *0.8654 & *0.8654 & *0.8763 & *0.8915 & *0.8813 & *0.8843 & *0.885 \\
    \cline{2-10}
    & \textbf{change vs. baseline} & -8.84\% & -6.34\% & -7.15\% & -7.71\% & -6.89\% & -6.93\% & -8.25\% & -8.57\% \\
    \midrule
    \texttt{fever} & \textbf{text-only} & 1.6366 & 1.5965 & 1.5683 & 1.4084 & 1.39 & 1.3764 & 1.306 & 1.3218 \\
    & \textbf{C-BN-text} & \textbf{*1.4653} & \textbf{*1.4837} & \textbf{*1.4503} & \textbf{1.3786} & \textbf{*1.3556} & \textbf{*1.3157} & \textbf{1.2763} & \textbf{*1.277} \\
    & \textbf{V-BN-text} & 1.7423 & 1.7089 & 1.701 & 1.5679 & 1.5602 & 1.5568 & 1.5109 & 1.5409 \\
    & \textbf{V-C-BN-text} & *1.511 & *1.5258 & *1.4959 & 1.4464 & 1.4325 & 1.3999 & 1.3855 & 1.4088 \\
    \cline{2-10}
    & \textbf{change vs. baseline} & -17.14\% & -11.28\% & -11.8\% & -2.98\% & -3.44\% & -6.07\% & -2.96\% & -4.48\% \\
    \midrule
    \texttt{mean} & \textbf{text-only} & 0.9524 & 0.9796 & 0.9756 & 0.9526 & 0.9518 & 0.9291 & 0.9215 & 0.9191 \\
    & \textbf{C-BN-text} & \textbf{*0.8409} & \textbf{*0.892} & \textbf{*0.8712} & \textbf{*0.8766} & \textbf{*0.8799} & \textbf{*0.867} & \textbf{*0.8598} & \textbf{*0.8599} \\
    & \textbf{V-BN-text} & *0.9172 & *0.9409 & *0.9366 & *0.9255 & *0.9265 & *0.8971 & *0.8907 & *0.8806 \\
    & \textbf{V-C-BN-text} & *0.8504 & *0.9018 & *0.8912 & *0.9049 & *0.9075 & *0.8942 & *0.8941 & *0.8953 \\
    \cline{2-10}
    & \textbf{change vs. baseline} & -11.15\% & -8.76\% & -10.44\% & -7.6\% & -7.19\% & -6.21\% & -6.16\% & -5.92\% \\
    \bottomrule
    \bottomrule
\end{tabular}}
\end{table*}

\begin{table*}[t]
  \caption{Brier scores ($\downdownarrows$) for our models on the \textit{not present, mentioned} subset across various training sizes. The best model per training size and per symptom is highlighted in \textbf{bold}. Cases where a model outperforms \textbf{text-only} significantly are indicated by \textbf{*} ($p < 0.05$ in a one-sided Wilcoxon signed-rank test over 20 seeds).}
  \label{tab:comparison_models_brier_npm}
\resizebox{\textwidth}{!}{
\begin{tabular}{llcccccccc}
    \toprule
    & & \multicolumn{8}{c}{Training size $n$} \\ \cmidrule{3-10}
    & & \textbf{100} & \textbf{187} & \textbf{350} & \textbf{654} & \textbf{1223} & \textbf{2287} & \textbf{4278} & \textbf{8000}\\
    \midrule
    \texttt{dysp} & \textbf{text-only} & 0.0358 & 0.028 & 0.0227 & 0.0188 & 0.0137 & 0.0135 & 0.0105 & 0.0053 \\
    & \textbf{C-BN-text} & *0.0339 & *0.0253 & *0.0215 & 0.0176 & 0.0134 & *0.0124 & 0.0096 & 0.0057 \\
    & \textbf{V-BN-text} & \textbf{*0.0302} & \textbf{*0.0243} & \textbf{*0.0211} & *0.0167 & 0.0137 & 0.0134 & 0.0105 & 0.0076 \\
    & \textbf{V-C-BN-text} & 0.0352 & *0.0258 & *0.0213 & \textbf{0.0166} & \textbf{0.0128} & \textbf{*0.0105} & \textbf{*0.0076} & \textbf{0.005} \\
    \cline{2-10}
    & \textbf{change vs. baseline} & -0.56\% & -0.37\% & -0.16\% & -0.22\% & -0.09\% & -0.3\% & -0.28\% & -0.03\% \\
    \midrule
    \texttt{cough} & \textbf{text-only} & 0.0929 & 0.0792 & 0.0656 & 0.0612 & 0.0473 & 0.0393 & 0.0239 & 0.0222 \\
    & \textbf{C-BN-text} & *0.0819 & *0.0725 & 0.0648 & 0.0571 & 0.044 & 0.0374 & 0.0253 & 0.0217 \\
    & \textbf{V-BN-text} & \textbf{*0.0771} & *0.0666 & *0.0588 & *0.0538 & *0.0413 & *0.0362 & 0.0241 & 0.0242 \\
    & \textbf{V-C-BN-text} & *0.0803 & \textbf{*0.064} & \textbf{*0.056} & \textbf{*0.0469} & \textbf{*0.0348} & \textbf{*0.0287} & \textbf{*0.0192} & \textbf{*0.0158} \\
    \cline{2-10}
    & \textbf{change vs. baseline} & -1.58\% & -1.52\% & -0.96\% & -1.43\% & -1.25\% & -1.05\% & -0.47\% & -0.64\% \\
    \midrule
    \texttt{pain} & \textbf{text-only} & 0.0293 & 0.032 & 0.0305 & 0.0275 & 0.0169 & 0.0127 & 0.0103 & 0.0066 \\
    & \textbf{C-BN-text} & *0.0246 & *0.0253 & *0.0208 & *0.0184 & *0.0134 & *0.0107 & 0.0089 & 0.007 \\
    & \textbf{V-BN-text} & \textbf{*0.0138} & \textbf{*0.0144} & \textbf{*0.0112} & \textbf{*0.009} & \textbf{*0.0055} & \textbf{*0.0043} & \textbf{*0.0035} & \textbf{*0.0025} \\
    & \textbf{V-C-BN-text} & *0.0264 & *0.0258 & *0.0203 & *0.0145 & *0.009 & *0.0063 & *0.0051 & *0.0036 \\
    \cline{2-10}
    & \textbf{change vs. baseline} & -1.55\% & -1.76\% & -1.94\% & -1.85\% & -1.13\% & -0.84\% & -0.67\% & -0.41\% \\
    \midrule
    \texttt{nasal} & \textbf{text-only} & 0.0584 & 0.0587 & 0.0514 & 0.0371 & 0.0299 & 0.0247 & 0.0155 & 0.0141 \\
    & \textbf{C-BN-text} & *0.0517 & *0.0528 & *0.0445 & *0.0336 & *0.0269 & *0.0215 & *0.0144 & *0.0129 \\
    & \textbf{V-BN-text} & \textbf{*0.0317} & \textbf{*0.0323} & \textbf{*0.0281} & \textbf{*0.0222} & \textbf{*0.0186} & *0.0172 & *0.0123 & *0.011 \\
    & \textbf{V-C-BN-text} & *0.0465 & *0.043 & *0.0354 & *0.0268 & *0.0197 & \textbf{*0.0144} & \textbf{*0.0099} & \textbf{*0.009} \\
    \cline{2-10}
    & \textbf{change vs. baseline} & -2.66\% & -2.63\% & -2.33\% & -1.49\% & -1.13\% & -1.03\% & -0.56\% & -0.52\% \\
    \midrule
    \texttt{fever} & \textbf{text-only} & 0.0929 & 0.1232 & 0.1025 & 0.1117 & 0.0691 & 0.0368 & 0.0318 & 0.019 \\
    & \textbf{C-BN-text} & *0.0728 & *0.0821 & *0.0697 & *0.0678 & *0.0467 & *0.0307 & *0.0244 & 0.0183 \\
    & \textbf{V-BN-text} & \textbf{*0.0278} & \textbf{*0.0361} & \textbf{*0.0263} & \textbf{*0.0329} & \textbf{*0.0195} & \textbf{*0.01} & \textbf{*0.0086} & \textbf{*0.0057} \\
    & \textbf{V-C-BN-text} & *0.0724 & *0.078 & *0.0623 & *0.0507 & *0.0301 & *0.0193 & *0.0142 & *0.0107 \\
    \cline{2-10}
    & \textbf{change vs. baseline} & -6.51\% & -8.71\% & -7.62\% & -7.88\% & -4.96\% & -2.68\% & -2.32\% & -1.34\% \\
    \midrule
    \texttt{mean} & \textbf{text-only} & 0.0619 & 0.0642 & 0.0545 & 0.0513 & 0.0354 & 0.0254 & 0.0184 & 0.0135 \\
    & \textbf{C-BN-text} & *0.053 & *0.0516 & *0.0443 & *0.0389 & *0.0289 & *0.0225 & *0.0165 & 0.0131 \\
    & \textbf{V-BN-text} & \textbf{*0.0361} & \textbf{*0.0347} & \textbf{*0.0291} & \textbf{*0.0269} & \textbf{*0.0197} & *0.0162 & *0.0118 & *0.0102 \\
    & \textbf{V-C-BN-text} & *0.0522 & *0.0473 & *0.0391 & *0.0311 & *0.0213 & \textbf{*0.0159} & \textbf{*0.0112} & \textbf{*0.0088} \\
    \cline{2-10}
    & \textbf{change vs. baseline} & -2.57\% & -2.95\% & -2.55\% & -2.43\% & -1.57\% & -0.95\% & -0.72\% & -0.47\% \\
    \bottomrule
    \bottomrule
\end{tabular}}
\end{table*}

\begin{table*}[t]
  \caption{Brier scores ($\downdownarrows$) for our models on the \textit{not present, not mentioned} subset across various training sizes. The best model per training size and per symptom is highlighted in \textbf{bold}. Cases where a model outperforms \textbf{text-only} significantly are indicated by \textbf{*} ($p < 0.05$ in a one-sided Wilcoxon signed-rank test over 20 seeds).}
  \label{tab:comparison_models_brier_npnm}
\resizebox{\textwidth}{!}{
\begin{tabular}{llcccccccc}
    \toprule
    & & \multicolumn{8}{c}{Training size $n$} \\ \cmidrule{3-10}
    & & \textbf{100} & \textbf{187} & \textbf{350} & \textbf{654} & \textbf{1223} & \textbf{2287} & \textbf{4278} & \textbf{8000}\\
    \midrule
    \texttt{dysp} & \textbf{text-only} & 0.0107 & 0.0091 & 0.0061 & 0.0049 & 0.0042 & 0.0044 & 0.0047 & 0.0039 \\
    & \textbf{C-BN-text} & 0.0111 & *0.0082 & 0.0061 & 0.0048 & 0.0042 & *0.004 & 0.0042 & 0.0039 \\
    & \textbf{V-BN-text} & \textbf{*0.0074} & \textbf{*0.0057} & \textbf{*0.0039} & \textbf{*0.003} & \textbf{*0.0027} & *0.0028 & *0.0031 & *0.0034 \\
    & \textbf{V-C-BN-text} & 0.0101 & *0.0073 & *0.0049 & *0.0038 & *0.0033 & \textbf{*0.0027} & \textbf{*0.0031} & \textbf{*0.0032} \\
    \cline{2-10}
    & \textbf{change vs. baseline} & -0.33\% & -0.34\% & -0.22\% & -0.19\% & -0.14\% & -0.17\% & -0.16\% & -0.07\% \\
    \midrule
    \texttt{cough} & \textbf{text-only} & \textbf{0.0158} & \textbf{0.0085} & \textbf{0.0058} & \textbf{0.0054} & 0.0057 & 0.0074 & 0.0061 & 0.0093 \\
    & \textbf{C-BN-text} & 0.018 & 0.0106 & 0.0093 & 0.007 & 0.0065 & 0.0076 & 0.0068 & 0.0087 \\
    & \textbf{V-BN-text} & 0.0186 & 0.0121 & 0.0098 & 0.0075 & 0.0062 & 0.007 & *0.0055 & *0.0082 \\
    & \textbf{V-C-BN-text} & 0.0173 & 0.0097 & 0.0078 & 0.0057 & \textbf{0.005} & \textbf{*0.0053} & \textbf{*0.0045} & \textbf{*0.0058} \\
    \cline{2-10}
    & \textbf{change vs. baseline} & +0.15\% & +0.12\% & +0.2\% & +0.03\% & -0.07\% & -0.21\% & -0.16\% & -0.35\% \\
    \midrule
    \texttt{pain} & \textbf{text-only} & 0.021 & 0.0164 & 0.0147 & 0.0131 & 0.0097 & 0.0098 & 0.0094 & 0.0081 \\
    & \textbf{C-BN-text} & 0.0196 & *0.015 & *0.013 & 0.0106 & 0.0091 & 0.0089 & 0.0086 & 0.0082 \\
    & \textbf{V-BN-text} & \textbf{*0.011} & \textbf{*0.0083} & \textbf{*0.0059} & \textbf{*0.0047} & \textbf{*0.0037} & \textbf{*0.0041} & \textbf{*0.0039} & \textbf{*0.0039} \\
    & \textbf{V-C-BN-text} & 0.0205 & *0.0147 & *0.0117 & *0.0081 & *0.0063 & *0.0058 & *0.0053 & *0.0049 \\
    \cline{2-10}
    & \textbf{change vs. baseline} & -1.01\% & -0.81\% & -0.88\% & -0.84\% & -0.6\% & -0.57\% & -0.55\% & -0.42\% \\
    \midrule
    \texttt{nasal} & \textbf{text-only} & 0.0112 & 0.0086 & 0.0066 & 0.0044 & 0.0033 & 0.0053 & 0.0028 & 0.0027 \\
    & \textbf{C-BN-text} & 0.0118 & 0.009 & 0.0068 & 0.0049 & 0.0038 & 0.0051 & 0.0033 & 0.0031 \\
    & \textbf{V-BN-text} & \textbf{*0.0083} & \textbf{*0.0068} & 0.006 & 0.0044 & 0.0034 & 0.0049 & 0.0038 & 0.0032 \\
    & \textbf{V-C-BN-text} & 0.0106 & *0.0074 & \textbf{0.0056} & \textbf{0.004} & \textbf{0.0026} & \textbf{*0.0028} & \textbf{*0.0018} & \textbf{*0.0014} \\
    \cline{2-10}
    & \textbf{change vs. baseline} & -0.29\% & -0.18\% & -0.1\% & -0.04\% & -0.07\% & -0.25\% & -0.1\% & -0.13\% \\
    \midrule
    \texttt{fever} & \textbf{text-only} & 0.0289 & 0.0293 & 0.0265 & 0.0527 & 0.05 & 0.0388 & 0.0427 & 0.029 \\
    & \textbf{C-BN-text} & 0.0367 & 0.0271 & 0.0278 & *0.0356 & *0.0359 & *0.0328 & *0.0331 & 0.0276 \\
    & \textbf{V-BN-text} & \textbf{*0.0096} & \textbf{*0.0088} & \textbf{*0.0072} & \textbf{*0.0163} & \textbf{*0.017} & \textbf{*0.0122} & \textbf{*0.0121} & \textbf{*0.0076} \\
    & \textbf{V-C-BN-text} & 0.0311 & *0.0223 & *0.0213 & *0.0249 & *0.0246 & *0.0211 & *0.0186 & *0.0138 \\
    \cline{2-10}
    & \textbf{change vs. baseline} & -1.92\% & -2.05\% & -1.93\% & -3.64\% & -3.3\% & -2.66\% & -3.06\% & -2.15\% \\
    \midrule
    \texttt{mean} & \textbf{text-only} & 0.0175 & 0.0144 & 0.0119 & 0.0161 & 0.0146 & 0.0131 & 0.0131 & 0.0106 \\
    & \textbf{C-BN-text} & 0.0194 & 0.014 & 0.0126 & *0.0126 & *0.0119 & *0.0117 & 0.0112 & 0.0103 \\
    & \textbf{V-BN-text} & \textbf{*0.011} & \textbf{*0.0084} & \textbf{*0.0065} & \textbf{*0.0072} & \textbf{*0.0066} & \textbf{*0.0062} & \textbf{*0.0057} & \textbf{*0.0053} \\
    & \textbf{V-C-BN-text} & 0.0179 & *0.0123 & *0.0103 & *0.0093 & *0.0083 & *0.0075 & *0.0067 & *0.0058 \\
    \cline{2-10}
    & \textbf{change vs. baseline} & -0.65\% & -0.6\% & -0.54\% & -0.89\% & -0.8\% & -0.7\% & -0.75\% & -0.53\% \\
    \bottomrule
    \bottomrule
\end{tabular}}
\end{table*}

\begin{table*}[t]
  \caption{The confidence of the \textbf{text-only} model, as measured by 1 - the normalized Shannon entropy \eqref{eq:confidence} on $\mathcal{X}_{test}$ over 20 seeds.}
  \label{tab:comparison_models_confidence}
  \resizebox{\textwidth}{!}{
\begin{tabular}{llcccccccc}
    \toprule
    & & \multicolumn{8}{c}{Training size $n$} \\ \cmidrule{3-10}
    & & \textbf{100} & \textbf{187} & \textbf{350} & \textbf{654} & \textbf{1223} & \textbf{2287} & \textbf{4278} & \textbf{8000}\\
    \midrule
    \texttt{dysp} & \textbf{text-only} & 0.7987 & 0.8466 & 0.8499 & 0.8794 & 0.8998 & 0.9025 & 0.9022 & 0.9237 \\
    \midrule
    \texttt{cough} & \textbf{text-only} & 0.6497 & 0.7263 & 0.7333 & 0.7833 & 0.8341 & 0.8637 & 0.8811 & 0.8914 \\
    \midrule
    \texttt{pain} & \textbf{text-only} & 0.6589 & 0.7571 & 0.7176 & 0.7494 & 0.7833 & 0.7898 & 0.8096 & 0.8279 \\
    \midrule
    \texttt{nasal} & \textbf{text-only} & 0.8164 & 0.872 & 0.876 & 0.8963 & 0.903 & 0.9058 & 0.9189 & 0.9243 \\
    \midrule
    \texttt{fever} & \textbf{text-only} & 0.6781 & 0.7484 & 0.7475 & 0.7855 & 0.8196 & 0.8534 & 0.8585 & 0.868 \\
    \bottomrule
    \bottomrule
\end{tabular}
}
\end{table*}

\subsection{Illustrative example} \label{sec:app_illustrative_example}

As explained in Section \ref{sec:pvm_subsets}, many of the cases contributing to the superior performance of the \textbf{V-C-BN-text} model over the \textbf{text-only} model are those where the symptom is present in the patient, but not present in the text. We zoom in on one of these cases to illustrate this point. 

Patient $8809$ in SimSUM has a common cold, and as a result of this, they suffer from a cough, nasal symptoms, a low fever, and pain. Furthermore, the tabular data record reveals that the patient visited the doctor's office during winter time, and stayed home for 9 days as a result of these symptoms. From the tabular evidence (\texttt{Common cold} = yes, \texttt{Season} = winter, \texttt{\#Days} = 9, all other tabular evidence = no), the Bayesian network predicts an 82.9\% chance for pain. The clinical note is as follows: 

\begin{small}
\begin{verbatim}
---
**History**
The patient presented with a constant cough persisting for the past week. They 
confirmed experiencing mild fever fluctuations, primarily in the evenings, 
describing the fever as low-grade. In addition, the patient reported nasal 
congestion and mild rhinorrhea, which started around the same time as the cough. 
The patient denied any episodes of dyspnea. They have been taking 
over-the-counter decongestants with minimal relief.

**Physical Examination**
Vital signs: Temperature: 37.6°C, Blood Pressure: 120/80 mmHg, Pulse: 78 bpm, 
Respiratory Rate: 16 breaths per  minute, and SpO2: 98% on room air. The patient 
appeared alert and in no acute distress. Upon auscultation, lungs were clear 
bilaterally with normal breath sounds and no wheezing or crackles. Nasal mucosa 
appeared slightly inflamed and there were no signs of significant throat 
erythema or exudates. Palpation of the cervical lymph nodes was unremarkable. 
Cardiac examination revealed regular rate and rhythm with no murmurs, rubs, or 
gallops. Abdomen was soft and non-tender with no apparent organomegaly.
---
\end{verbatim}
\end{small}

This note does not mention pain anywhere. By training on similar examples, the neural text classifier has learned that there is a very low chance of pain in this patient (since most notes in the training set that are labeled with pain, do indeed mention pain somewhere in the note). As a result, the text classifier predicts only a 4.9\% chance for pain. 

By combining the BN's prediction with the neural text classifier's prediction using the \textbf{V-C-BN-text} model, we arrive at a probability of 40.1\% for pain, which is higher than before and closer to the true label.\footnote{These results are attained for seed 2014 on training size 187.}

\subsection{Handling shifts in text data distribution}
\label{sec:app_data_shift_results}

As explained in Section \ref{sec:data_shift}, we used our original \textbf{BN-only} and \textbf{text-only} classifiers, as well as the original consistency nodes (which were trained using the original train set $\mathcal{X}_{train}$ with non-manipulated notes) to evaluate performance on the new test set $\mathcal{X}_{test}^*$ with manipulated notes. Tables \ref{tab:data_shift_average_precision_all} and \ref{tab:data_shift_Brier_all} show the results for all symptoms. The \textbf{V-C-BN-text} model significantly improves over the \textbf{text-only} baseline on both metrics for almost all training sizes and symptoms. These manipulated notes are released on our Github repository, along with the rest of the code, at \url{https://github.com/AdrickTench/patient-level-IE}.

\begin{table*}[t]
  \caption{Average precision ($\upuparrows$) for the predictions of our models over the test set $\mathcal{X}_{test}^*$ containing \textbf{manipulated text notes}. The best model per training size and per symptom is highlighted in \textbf{bold}. Cases where a model outperforms \textbf{text-only} significantly are indicated by \textbf{*} ($p < 0.05$ in a one-sided Wilcoxon signed-rank test over 20 seeds).}
  \label{tab:data_shift_average_precision_all}
\resizebox{\textwidth}{!}{
\begin{tabular}{llcccccccc}
    \toprule
    & & \multicolumn{8}{c}{Training size $n$} \\ \cmidrule{3-10}
    & & \textbf{100} & \textbf{187} & \textbf{350} & \textbf{654} & \textbf{1223} & \textbf{2287} & \textbf{4278} & \textbf{8000}\\
    \midrule
    \texttt{dysp} & \textbf{text-only} & 0.8762 & 0.9068 & 0.9206 & 0.9342 & 0.943 & 0.9519 & 0.9552 & 0.961 \\
    & \textbf{C-BN-text} & *0.8891 & *0.9122 & *0.9245 & *0.9366 & *0.9455 & 0.9515 & 0.9551 & 0.9616 \\
    & \textbf{V-BN-text} & *0.8905 & *0.9114 & 0.9204 & *0.94 & \textbf{*0.948} & *0.9553 & *0.9587 & *0.9643 \\
    & \textbf{V-C-BN-text} & \textbf{*0.8915} & \textbf{*0.9172} & \textbf{*0.9278} & \textbf{*0.9404} & *0.9478 & \textbf{*0.9555} & \textbf{*0.9592} & \textbf{*0.9645} \\
    \cline{2-10}
    & \textbf{change vs. baseline} & +1.53\% & +1.04\% & +0.72\% & +0.62\% & +0.5\% & +0.36\% & +0.41\% & +0.35\% \\
    \midrule
    \texttt{cough} & \textbf{text-only} & 0.825 & 0.8651 & 0.881 & 0.8992 & 0.9144 & 0.9251 & 0.9328 & 0.9345 \\
    & \textbf{C-BN-text} & *0.8515 & *0.8847 & *0.902 & *0.9146 & *0.9274 & *0.9353 & *0.9428 & *0.9431 \\
    & \textbf{V-BN-text} & *0.8574 & \textbf{*0.8942} & \textbf{*0.9106} & \textbf{*0.9254} & \textbf{*0.938} & \textbf{*0.9451} & \textbf{*0.9519} & \textbf{*0.955} \\
    & \textbf{V-C-BN-text} & \textbf{*0.8586} & *0.8928 & *0.9071 & *0.9193 & *0.9313 & *0.9394 & *0.9471 & *0.95 \\
    \cline{2-10}
    & \textbf{change vs. baseline} & +3.35\% & +2.92\% & +2.97\% & +2.63\% & +2.36\% & +2.0\% & +1.91\% & +2.05\% \\
    \midrule
    \texttt{pain} & \textbf{text-only} & 0.5357 & 0.6195 & 0.6575 & 0.6946 & 0.722 & 0.7349 & 0.7422 & 0.7516 \\
    & \textbf{C-BN-text} & \textbf{*0.5467} & \textbf{*0.6262} & \textbf{*0.669} & *0.7031 & *0.7285 & *0.7404 & *0.7493 & *0.7595 \\
    & \textbf{V-BN-text} & 0.5176 & 0.6106 & 0.6567 & \textbf{*0.714} & \textbf{*0.7473} & \textbf{*0.764} & \textbf{*0.775} & \textbf{*0.7868} \\
    & \textbf{V-C-BN-text} & 0.538 & 0.6228 & *0.6632 & *0.7065 & *0.737 & *0.7528 & *0.7613 & *0.7723 \\
    \cline{2-10}
    & \textbf{change vs. baseline} & +1.1\% & +0.67\% & +1.15\% & +1.94\% & +2.53\% & +2.91\% & +3.27\% & +3.52\% \\
    \midrule
    \texttt{nasal} & \textbf{text-only} & 0.8788 & 0.8869 & 0.9013 & 0.9072 & 0.9117 & 0.9097 & 0.9101 & 0.9034 \\
    & \textbf{C-BN-text} & *0.9032 & *0.9073 & *0.9205 & *0.927 & *0.9308 & *0.9263 & *0.9288 & *0.9238 \\
    & \textbf{V-BN-text} & \textbf{*0.9057} & \textbf{*0.9116} & \textbf{*0.9229} & \textbf{*0.9307} & \textbf{*0.9369} & \textbf{*0.9371} & \textbf{*0.9388} & \textbf{*0.9337} \\
    & \textbf{V-C-BN-text} & *0.8986 & *0.9052 & *0.919 & *0.9238 & *0.9288 & *0.928 & *0.9296 & *0.9231 \\
    \cline{2-10}
    & \textbf{change vs. baseline} & +2.69\% & +2.47\% & +2.16\% & +2.35\% & +2.52\% & +2.74\% & +2.87\% & +3.03\% \\
    \midrule
    \texttt{fever} & \textbf{text-only} & 0.6023 & 0.6524 & 0.6875 & 0.7361 & 0.7714 & 0.7909 & 0.8072 & 0.8161 \\
    & \textbf{C-BN-text} & 0.5975 & *0.6562 & *0.6943 & *0.7424 & *0.7761 & *0.7981 & *0.8132 & *0.8243 \\
    & \textbf{V-BN-text} & \textbf{0.6047} & \textbf{*0.6782} & \textbf{*0.7202} & \textbf{*0.7746} & \textbf{*0.8086} & \textbf{*0.8269} & \textbf{*0.8439} & \textbf{*0.8543} \\
    & \textbf{V-C-BN-text} & 0.6034 & *0.6671 & *0.7044 & *0.7572 & *0.7915 & *0.8104 & *0.829 & *0.839 \\
    \cline{2-10}
    & \textbf{change vs. baseline} & +0.23\% & +2.58\% & +3.27\% & +3.86\% & +3.72\% & +3.6\% & +3.67\% & +3.83\% \\
    \midrule
    \texttt{mean} & \textbf{text-only} & 0.7436 & 0.7861 & 0.8096 & 0.8342 & 0.8525 & 0.8625 & 0.8695 & 0.8733 \\
    & \textbf{C-BN-text} & *0.7576 & *0.7973 & *0.822 & *0.8447 & *0.8617 & *0.8703 & *0.8778 & *0.8825 \\
    & \textbf{V-BN-text} & *0.7552 & \textbf{*0.8012} & \textbf{*0.8262} & \textbf{*0.857} & \textbf{*0.8758} & \textbf{*0.8857} & \textbf{*0.8937} & \textbf{*0.8988} \\
    & \textbf{V-C-BN-text} & \textbf{*0.758} & *0.801 & *0.8243 & *0.8495 & *0.8673 & *0.8772 & *0.8853 & *0.8898 \\
    \cline{2-10}
    & \textbf{change vs. baseline} & +1.44\% & +1.51\% & +1.66\% & +2.27\% & +2.32\% & +2.32\% & +2.42\% & +2.55\% \\
    \bottomrule
    \bottomrule
\end{tabular}}
\end{table*}

\begin{table*}[t]
  \caption{Brier scores ($\downdownarrows$) for the predictions of our models over the test set $\mathcal{X}_{test}^*$ containing \textbf{manipulated text notes}. The best model per training size and per symptom is highlighted in \textbf{bold}. Cases where a model outperforms \textbf{text-only} significantly are indicated by \textbf{*} ($p < 0.05$ in a one-sided Wilcoxon signed-rank test over 20 seeds).}
  \label{tab:data_shift_Brier_all}
\resizebox{\textwidth}{!}{
\begin{tabular}{llcccccccc}
    \toprule
    & & \multicolumn{8}{c}{Training size $n$} \\ \cmidrule{3-10}
    & & \textbf{100} & \textbf{187} & \textbf{350} & \textbf{654} & \textbf{1223} & \textbf{2287} & \textbf{4278} & \textbf{8000}\\
    \midrule
    \texttt{dysp} & \textbf{text-only} & 0.062 & 0.051 & 0.0452 & 0.0419 & 0.0377 & 0.0337 & 0.032 & 0.028 \\
    & \textbf{C-BN-text} & 0.063 & *0.0497 & *0.0443 & *0.0401 & *0.0365 & *0.0326 & *0.031 & *0.0275 \\
    & \textbf{V-BN-text} & 0.0606 & *0.0492 & 0.0441 & 0.0406 & 0.0373 & 0.0336 & 0.0326 & 0.0305 \\
    & \textbf{V-C-BN-text} & \textbf{*0.058} & \textbf{*0.0469} & \textbf{*0.0419} & \textbf{*0.038} & \textbf{*0.0349} & \textbf{*0.0312} & \textbf{*0.0296} & \textbf{0.0274} \\
    \cline{2-10}
    & \textbf{change vs. baseline} & -0.4\% & -0.41\% & -0.34\% & -0.39\% & -0.28\% & -0.25\% & -0.24\% & -0.05\% \\
    \midrule
    \texttt{cough} & \textbf{text-only} & 0.137 & 0.1212 & 0.1141 & 0.105 & 0.0973 & 0.0889 & 0.087 & 0.0847 \\
    & \textbf{C-BN-text} & *0.12 & *0.1084 & *0.0992 & *0.094 & *0.0872 & *0.081 & *0.0786 & *0.0778 \\
    & \textbf{V-BN-text} & *0.1213 & \textbf{*0.1039} & \textbf{*0.0958} & \textbf{*0.0886} & \textbf{*0.0818} & \textbf{*0.0754} & \textbf{*0.0721} & \textbf{*0.0694} \\
    & \textbf{V-C-BN-text} & \textbf{*0.1173} & *0.105 & *0.0965 & *0.091 & *0.0838 & *0.0783 & *0.0756 & *0.0743 \\
    \cline{2-10}
    & \textbf{change vs. baseline} & -1.97\% & -1.73\% & -1.82\% & -1.64\% & -1.55\% & -1.35\% & -1.49\% & -1.53\% \\
    \midrule
    \texttt{pain} & \textbf{text-only} & 0.0955 & 0.0879 & 0.0793 & 0.0753 & 0.0698 & 0.0665 & 0.0657 & 0.0621 \\
    & \textbf{C-BN-text} & 0.0982 & 0.0867 & 0.082 & 0.0754 & 0.0703 & 0.0679 & 0.0662 & 0.0628 \\
    & \textbf{V-BN-text} & 0.1043 & 0.0933 & 0.0869 & 0.0773 & 0.0733 & 0.0694 & 0.067 & 0.0639 \\
    & \textbf{V-C-BN-text} & \textbf{0.0949} & \textbf{*0.0853} & \textbf{0.0789} & \textbf{*0.072} & \textbf{0.0683} & \textbf{0.0661} & \textbf{0.0641} & \textbf{0.0618} \\
    \cline{2-10}
    & \textbf{change vs. baseline} & -0.05\% & -0.26\% & -0.04\% & -0.33\% & -0.15\% & -0.04\% & -0.16\% & -0.03\% \\
    \midrule
    \texttt{nasal} & \textbf{text-only} & 0.0905 & 0.0847 & 0.0767 & 0.0751 & 0.0708 & 0.0682 & 0.0677 & 0.0688 \\
    & \textbf{C-BN-text} & *0.0869 & *0.0817 & *0.0737 & *0.0713 & \textbf{*0.0678} & \textbf{*0.0657} & \textbf{*0.065} & \textbf{*0.0661} \\
    & \textbf{V-BN-text} & 0.0915 & 0.0832 & 0.0761 & 0.0758 & 0.0712 & 0.0679 & 0.0673 & *0.0668 \\
    & \textbf{V-C-BN-text} & \textbf{*0.0834} & \textbf{*0.0798} & \textbf{*0.0727} & \textbf{*0.071} & *0.069 & 0.068 & 0.0683 & 0.0694 \\
    \cline{2-10}
    & \textbf{change vs. baseline} & -0.71\% & -0.48\% & -0.4\% & -0.42\% & -0.3\% & -0.24\% & -0.27\% & -0.27\% \\
    \midrule
    \texttt{fever} & \textbf{text-only} & 0.3071 & 0.2887 & 0.2695 & 0.2613 & 0.2388 & 0.2231 & 0.2134 & 0.2041 \\
    & \textbf{C-BN-text} & 0.3029 & *0.2743 & *0.2612 & *0.2401 & *0.2222 & *0.2102 & *0.2012 & *0.196 \\
    & \textbf{V-BN-text} & 0.3233 & 0.2878 & 0.2736 & *0.2437 & *0.2251 & 0.2171 & 0.2051 & 0.2036 \\
    & \textbf{V-C-BN-text} & \textbf{*0.294} & \textbf{*0.2671} & \textbf{*0.2515} & \textbf{*0.2306} & \textbf{*0.2131} & \textbf{*0.2034} & \textbf{*0.1949} & \textbf{*0.1929} \\
    \cline{2-10}
    & \textbf{change vs. baseline} & -1.31\% & -2.16\% & -1.8\% & -3.07\% & -2.57\% & -1.97\% & -1.85\% & -1.12\% \\
    \midrule
    \texttt{mean} & \textbf{text-only} & 0.1384 & 0.1267 & 0.117 & 0.1117 & 0.1028 & 0.0961 & 0.0932 & 0.0895 \\
    & \textbf{C-BN-text} & *0.1342 & *0.1202 & *0.1121 & *0.1042 & *0.0968 & *0.0915 & *0.0884 & *0.0861 \\
    & \textbf{V-BN-text} & 0.1402 & *0.1235 & *0.1153 & *0.1052 & *0.0977 & *0.0927 & *0.0888 & *0.0868 \\
    & \textbf{V-C-BN-text} & \textbf{*0.1295} & \textbf{*0.1168} & \textbf{*0.1083} & \textbf{*0.1005} & \textbf{*0.0938} & \textbf{*0.0894} & \textbf{*0.0865} & \textbf{*0.0852} \\
    \cline{2-10}
    & \textbf{change vs. baseline} & -0.89\% & -0.99\% & -0.87\% & -1.12\% & -0.9\% & -0.67\% & -0.67\% & -0.44\% \\
    \bottomrule
    \bottomrule
\end{tabular}}
\end{table*}

\end{appendices}

\end{document}